\documentclass[journal]{IEEEtran}
\ifCLASSINFOpdf
\else
\fi

\usepackage{xcolor}
\usepackage{algorithm}
\usepackage{algorithmic}

\newcommand{\otcxx}[1]{\textcolor{black}{#1}}
\newcommand{\otc}[1]{\textcolor{black}{#1}}
\newcommand{\ccc}[1]{\textcolor{black}{#1}}

\newcommand{\br}[1]{\textcolor{black}{#1}}

\newcommand{\cc}[1]{\textcolor{black}{#1}}
\newcommand{\brtwo}[1]{\textcolor{black}{#1}}

\newcommand{\TIPanswer}[1]{\textcolor{black}{#1}}

\newcommand{\TIPnewanswer}[1]{\textcolor{black}{#1}}
\usepackage{multirow}
\usepackage{makecell}
\usepackage{graphicx}
\graphicspath{ {./Images/} {./} {../} {/}}
\usepackage{url}

\usepackage{amsfonts}
\usepackage{amsmath}

\DeclareMathOperator*{\argmin}{arg\,min}

\newcommand{\omitme}[1]{}

\begin{document}

\title{Inspirational Adversarial Image Generation}

\author{Baptiste Rozi\`ere, \and Morgane Riviere, \and 
        Olivier Teytaud, \and 
        Jeremy Rapin, \and
        Yann LeCun, \and
        Camille Couprie}

\maketitle

\begin{abstract}
The task of image generation started receiving some attention from artists and designers, providing inspiration for new creations. However, exploiting the results of deep generative models such as Generative Adversarial Networks can be long and tedious given the lack of existing tools. 
In this work, we propose a simple strategy to inspire creators with new generations learned from a dataset of their choice, while providing some control over the output. We design a simple optimization method to find the optimal latent parameters corresponding to the closest generation to any input inspirational image.
Specifically, we allow the generation given an inspirational image of the user's choosing by performing several optimization steps to recover optimal parameters from the model's latent space. We tested several exploration methods from  classical gradient descents to gradient-free optimizers.
Many gradient-free optimizers just need comparisons (better/worse than another image), so they can even be used without numerical criterion nor inspirational image, only with human preferences. Thus, by iterating on one's preferences we can make robust facial composite or fashion generation algorithms. 
Our results on four datasets of faces, fashion images, and textures show that satisfactory images are effectively retrieved in most cases.

\end{abstract}

\begin{IEEEkeywords}
Optimization, \and Generative adversarial networks, \and Similarity search
\end{IEEEkeywords}

\IEEEpeerreviewmaketitle

%---------------------
\section{Introduction}
%---------------------
\begin{figure}[htb]
\begin{center}
\includegraphics[width=1\linewidth]{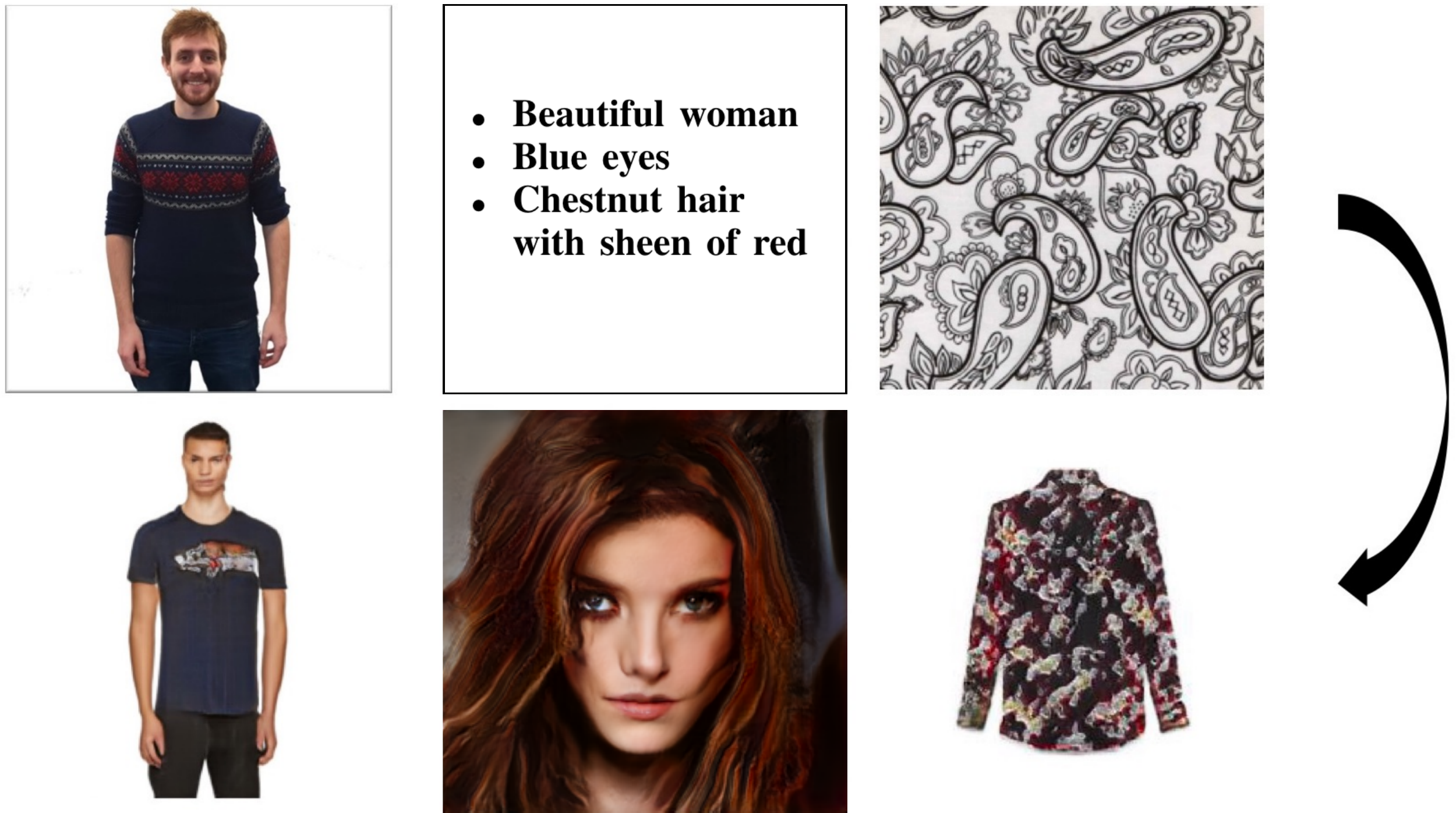}
\end{center}
\caption{Generative networks may be a source of inspiration for designers and artists, but the current approaches lack  control over generations. Our approach allows to discover generations (bottom line) from trained networks and either (i) an inspirational image (top line) or (ii) preferences among presented images provided by a user.}
\label{fig:teaser}
\end{figure}

% Motivate problem: utility, challenges
Generative models, and in particular adversarial ones \cite{goodfellow2014generative}, are becoming prevalent in computer vision as they enable enhancing artistic creation \cite{elgammal2017,Zhu2017cycleGANs,Park2019spade}, inspire designers \cite{sbai2018design,ZhuUrtasun2017iccvPrada}, or prove usefulness in semi-supervised learning \cite{Donahue2016Bigan,frid2018gan,nie2017medical}.
\cc{In this image generation context, }GAN models typically take a random vector as input and output an image. Due to the high dimensionality of their latent space, we know little about what kind of picture we will get. To reduce the burden on creative people compelled to brute-force GAN's generations and cherry pick nice examples, we propose to generate at test time either \emph{image}-inspired or \emph{preference}-based generations.
 \ccc{
 Many approaches such as style transfer \cite{Gatys2016ImageStyleTransfer}, or GANs conditioned on semantic information \cite{Park2019spade} \TIPanswer{or text
 \cite{reed2016generative}} make the choice of incorporating constraints in the training procedure. In contrast, we explore  the retrieval of interesting generations that exist in the latent space of a trained GAN. There are \TIPanswer{three} main advantages to this approach: first, a fast recovery of results is possible without the requirement of training the model again. \TIPanswer{Second, there is a guarantee that the obtained image has not been altered by the inspiration image, and third, it allows to use completely unexpected inspiration targets that are not present in training data.} Being able to control the output of GANs is a first step toward numerous applications.
 The primary motivation of our work is to provide a tool for artists looking for inspiration to generate interesting images given a picture, a description, or even just an imprecise idea of what they want to obtain, \cc{see Figure~\ref{fig:teaser}.} For instance, \ccc{some} artists admit spending many hours randomly browsing GAN generations \cite{Spratt2018Klingemann}.
 It may also be useful to designers who want to invent new clothes similar to successful items of past collections, or a given piece of textile \cite{sbai2018design}. The design of new artificial textures or avatars \cite{Wolf2017avatar} for games may be an application as well.  Finally, fake item generations could be used in various other contexts, from photo editing software (\otc{e.g. inpainting}  \cite{Ulyanov2018deep,Yeh_2017_CVPR}, \otc{super-resolution} \cite{srinspir,roziere2020tarsier}) to facial \otc{composites \cite{tog}} or anonymization \cite{Gafni_2019_ICCV}.}\TIPanswer{ The choice of providing control using spatial information or hollistic feedbacks rather than text is motivated by the better results this richer source can provide. For instance, \cite{solomon2009new} mentions that holistic methods, in which users enter global comparisons between faces, are more reliable than methods based on detailed descriptions, when trying to render human memories.}

We present two options for latent space exploration: 
\begin{itemize}
	\item Given an inspirational image and a trained model, we recover via \ccc{numerical optimization (gradient-based or gradient-free)} the input vector that minimizes a reconstruction criterion defined between the given image and the generation. 
	\item \br{Alternatively, we can put humans in the loop and, at each iteration, we base our criterion on human \ccc{image} preferences. We converge to an image similar to the target image in very few iterations (i.e. 6).} 
\end{itemize} Despite a highly non-convex optimization, we show that it is possible to obtain related generations.

\medskip

Our contributions include
\begin{itemize}

\item \ccc{Defining and validating} a criterion of generation retrieval given a target image. \ccc{The dimensions of our exploration include a loss on image intensities, a perceptual loss using pre-trained features,  a penalization of the norm in the latent space and finally, optionally  on the discriminator's output.} \TIPanswer{We note that combinations of pixel and feature level losses exist in previous works including \cite{ledig2017photo,zamir2019learning,jackson2019style}. Our loss combination is novel in the context of image retrieval.}
\item \otc{Validating the use of evolutionary algorithms both for incorporating humans preferences} \ccc{in GAN generations} (as in our proposed HEVOL, \ccc{extending \cite{tog})} \otc{and for finding better  optima from a human point of view.}
\item \ccc{Validating/justifying the use of LBFGS in numerical terms, as opposed to methods classically used in machine learning such as stochastic gradient descent or Adam.}
\item \ccc{Showcasing different applications and comparing different optimizers }\otc{on various datasets, from classical faces (Celeba-HQ) for facial composite applications to FashionGEN and textures (DTD),} \ccc{for  garment image designs generation}.
\end{itemize}

It is worth noticing that our procedure is done at test time and, compared to other methods \cite{imtoim,baubau}, does not require any additional update of the model's weights or other models. The control provided by our approach is both simple, intuitive, and computationally efficient. After introducing the related work in Section II, we present our approach to retrieve inspired generations from trained GANs in Section III, and experiments that include the comparison of various similarity criteria, several gradient-based and gradient-free optimization approaches on diverse datasets in Sections IV and V. 

%---------------------
\section{Related work}
%---------------------

\ccc{In this section, we %after motivating our choice of GAN setting, we %detail some composite GAN methods related to our decoupled GAN, and 
discuss related works exploring the latent space of GANs, and review relevant user-based image generation approaches.}

\textbf{Latent space exploration of GANs.}
The exploration of the latent space of GANs was popularized by the DCGAN work presenting latent space interpolations and arithmetic operation results \cite{radford2015unsupervised}.
Learning a mapping projecting data back in the latent space of GANs has been studied in the context of bi-directional GANs \cite{Donahue2016Bigan}, with an emphasis on a utility in semi-supervised learning. Similarly, image generation may improve zero shot learning  tasks \cite{zhu2018generative}. In Fader Networks \cite{lample2017fader}, image manipulation is made possible by learning an image representation disentangled from its attributes with adversarial training. 
Recently, the Style-based generator architecture of Karras et al. \cite{karras2018style} improved the coherency of the generator's internal representation by creating an intermediate latent space and enforcing feature statistics proximity between neighbor codes. The criterion of feature similarity is borrowed from Texture networks \cite{Ulyanov2016textureNet}. Other studies explored the latent space of GANs in order to improve the quality of the generated images according to an image quality assessor~\cite{roziere2020tarsier,roziere2020_evolgan}. 

A number of works focus on neural visualization of trained networks for image classification  \cite{zeiler2014visualizing,DosovitskiyNIPS2016,Nguyen2016preferred}. 
A somehow related task lies in membership inference, where the goal is to determine if an image has been seen during training \cite{sablayrolles2018deja,koh2017understanding}. 
The notion of feature similarity in image generation is widely employed, for instance in  style transfer \cite{Gatys2016ImageStyleTransfer}, and generation quality assessment \cite{Zhang2018unreasonable}. 
The most similar work in spirit to our image inspiration strategy is the Inference-via-optimization approach defined to evaluate the severity of mode collapse in Unrolled GANs \cite{unrolledgan}. We can also cite the approach of \cite{fairlatent} that also matches vectors in the latent space of GANs with precise pictures at training time using a Nesterov gradient. 
\ccc{Finally, \cite{invert} studies ``inversion of the generator" by gradient descent.} 

In contrast to these works, we carefully design an appropriate similarity metric based on semantic features that also exploits the discriminator. Moreover, we benchmark various search strategies to obtain better generations including gradient-free approaches which do not always need a target image.

\textbf{Interactive Image Generation.}
\ccc{A number of works use user inputs for driving the generations, for instance user preferences \cite{tog}, or regions to modify \cite{baubau,Zhu2016manip,Park2019spade}. 
 Bontrager et al. \cite{tog} is the first to suggest the use of evolutionary approaches to perform interactive generation of images of faces and shoes. We extend this work by also studying the behavior of evolutionary approaches given targets, define a similarity criterion explaining the better quality of our generated images, and assess the optimality of results with respect to this criterion as well as human perception.} 
 
 \TIPanswer{Another possibility to provide control on generations could be using text inputs. 
 The work of Zhu et al. \cite{ZhuUrtasun2017iccvPrada} is among the first successful approaches to generate fashion images with GANs, conditioned on text and segmentation masks.
The work of Rostamzadeh et al. \cite{fashionGen2018} introduced the Fashion-Gen fashion image dataset, with progressive growing of GAN generations conditioned by text inputs. The produced generations are mostly uniform clothing items, because most images from the dataset display single color garments, and text descriptions are limited. HD-GANs \cite{zhang2018photographic} achieves correct image quality on fine-grained image datasets, (e.g. birds, flowers), where objects can easily be described by texts. Text is less informative than spatial information, at least for faces \cite{nasir2019text2facegan} or texture images.
}
 
\ccc{Several approaches incorporate user interaction in the generative process by the mean of paint-brush strokes to guide GAN \cite{Zhu2016manip,Park2019spade} of hybrid VAE-GAN generations \cite{Brock2016neural}. In an image editing context, the work of Yeh et al. \cite{Yeh_2017_CVPR} deals with inpainting by iteratively updating the latent code of generations.
\cite{baubau} suggests to edit images in three steps. First, a latent code is computed from an original image. Second, this code is edited to produce a satisfactory result in a region defined by a user. Finally, the generator is updated to guarantee fidelity to the original image in the unedited regions. All the previously mentioned works focus on particular applications and are orthogonal to our work.}

%--------------------
\section{Approach}
%--------------------

\label{sec:approach}

\subsection{Background}

GAN models are built as follows: given two networks, a generator $G$ and a discriminator $D$, the discriminator is trained to differentiate real data from fake ones, while $G$ \cc{receives feedback on how to update its weights to fool $D$.} 

\textbf{Progressive growing of GANs (PGAN).} 
\ccc{We rely on a} PGAN architecture \cite{karras2017progressive} in order to reach high resolutions. \ccc{This} method trains both networks partially, one resolution at a time. In other words, it starts with two small networks trained on $4\times4$ images; once the results are satisfying, the layers corresponding to the $8\times8$ resolution are added and the training continues. This method goes on until the desired resolution is reached, \ccc{for instance up to $1024\times1024$}.

The progressive growing method happens to be very stable: even in high resolution, we did not notice any mode collapse in any of our training sessions. Besides, we show that it gives convincing results even with small datasets ($\approx 4000$ images). Our implementation of Progressive Growing and the inspiration method described in this paper is open-sourced\footnote{\url{https://github.com/facebookresearch/pytorch_GAN_zoo}}.

\textbf{Conditional generation with GANs.} 
Odena \& al. \cite{Odena2017conditionalGAN} proposed a simple method for enforcing label conditioning at training time on GANs: labels are encoded as a one-hot vector concatenated to the continuous generator's latent space, the generator is trained to generate data coherent with their given labels and the discriminator is trained to recognize them.

\subsection{Multi-class conditioning}

\cc{We trained PGAN models without any label conditioning in the case of face datasets, and used available labels to improve results in the other employed datasets. When using conditioning, }
we apply a variation of AC-GAN \cite{Odena2017conditionalGAN}. If a class $c$ has $k_c$ possible values, then it can be encoded in a one-hot vector $(c_1, \dots, c_{k_c})$ with  $c_i = 1$ codes for label $i$. 
When $C>1$ groups of classes exist, we concatenate the encoding vectors of all classes to get a label vector  \TIPanswer{
$\hat{c}=((c^1_1,\dots,c^1_{k_1}), 
(c^2_1,\dots,c^2_{k_2}),\dots,(c^C_1,\dots,c^C_{k_C}))$.}
This label vector is then fed to the generator with the latent input noise $z$.
The classification loss becomes:
{
\begin{equation}
   \mathcal{L}_{\scriptsize{\mbox{class}}}(\hat{c}, x) = - \TIPanswer{\sum_{c\TIPanswer{=1}}^{\TIPanswer{C}}} \sum_{i=1}^{k_c} \log \left(\TIPanswer{\frac{e^{D_{\TIPanswer{c^c_i}}(x)}}{\sum_{q=1}^{k_c} e^{D_{{c^c_q}}(x)}}}\right).
\end{equation}}
The discriminator $D$ {is trained to} minimize
\begin{equation}
\mathcal{L}_D = \mathcal{L}_{PGAN(D)} + \lambda_D^c \mathcal{L}_{\scriptsize{\mbox{class}}}(\hat{c}_{real}, x_{real}),
\end{equation}
while the generator $G$ {is trained to} minimize
{\begin{equation}
\TIPanswer{\mathcal{L}_{G}=}  \mathcal{L}_{PGAN(G)} + \lambda_G^c \mathcal{L}_{\scriptsize{\mbox{class}}}(\hat{c}_{noise}, z),\end{equation}}
{where} $\mathcal{L}_{PGAN(D)}$ and $\mathcal{L}_{PGAN(G)}$ \ccc{are} the loss penalties of Progressive Growing, $\hat{c}_{real}$ is the label vector of the input image $x_{real}$ and $\hat{c}_{noise}$ the random label vector fed to $G$ along with $z$. \cc{The scalars $\lambda_G^c$ and $\lambda_D^c$ are scale parameters.}

\subsection{Image-inspired image generation}

\begin{figure}[htb]
\begin{center}
\includegraphics[width=1\linewidth]{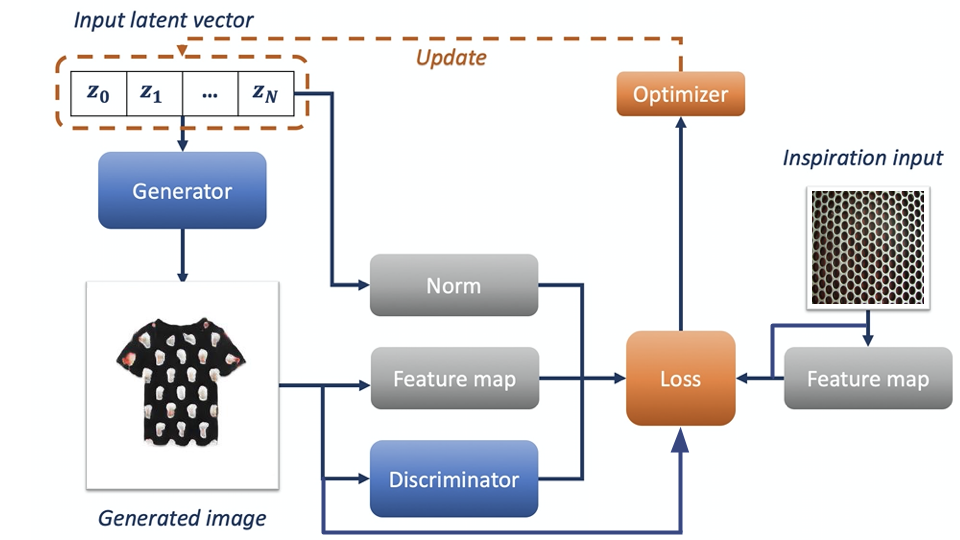}
\end{center}
\caption{Our inspiration based approach. On a trained GAN, we perform a search on the latent vector \ccc{$z$} fed to the generator in order to find the \br{best generation for a given reference image}. \otc{The distance/loss can use features, simple Euclidean norm or use the discriminator.}}
\label{fig:insp}
\end{figure}

\cc{In this section, we design an optimization criterion to retrieve GAN samples inspired by an input image. Our pipeline is depicted in Figure \ref{fig:insp}.}

\label{sec:approach_insp} 

The different directions of the latent space of a GAN model usually do not make sense from a human point of view. We wanted to know if it was possible to explore this space to perform what we call an ``inspirational generation''. In other words, given a reference image $I$ we would like to force the generator to produce an output as similar as possible to $I$. 

To do so, we consider performing a gradient descent on the noise vector $z$ fed to the generator. This idea raises two issues:
\begin{itemize}
    \item How do we define an objective criterion for image similarity?
    \item Can we converge to an optimal vector?
\end{itemize}

\subsubsection{Similarity}

We first need a similarity measure between images. Therefore, we tried two different approaches.

\paragraph{Pixel-based similarity}

The simplest way to estimate the similarity between two pictures is to perform their difference pixel-wise. This give us a similarity criterion $\mathcal{C}_L$:
\begin{equation}
    \mathcal{C}_L(z, I) = \frac{1}{N_p} \sum_{p \in pixels} \| G(z)_p - I_p \|^2,
\end{equation}
\cc{where $N_p$ is the number of pixels of $I$ \TIPanswer{and $I_p$ the $p^{th}$ pixel of $I$.}}
This loss is not invariant by translation, rotation or illumination change. Also, we do not expect to capture abstract concepts (gender, posture, \ccc{complex texture patterns,} etc.) with it.

\paragraph{Feature-based similarity}
We found that a variation of the style criterion used in \cite{huang2017styletransfer} performs well for nearest neighbor search in several considered datasets. This criterion $\mathcal{C}_S$ is defined as
\begin{multline}
\mathcal{C}_S(z,I) = \sum_{i} \|\mu(\phi_i(G(z))) - \mu(\phi_i(\TIPanswer{I})) \|^2 \\
+ \sum_{i} \|\sigma(\phi_i(G(z)))^2 - \sigma(\phi_i(\TIPanswer{I}))^2 \|,
\end{multline}
where:
\begin{itemize}
    \item $\phi_i$ represents the \TIPanswer{output of the  $i^{th}$} layer of a fully convolutional neural network trained on \br{ImageNet}, 
    \item \TIPanswer{$i$ ranges over a list of layer indices (see Appendix),}
    \item $\mu$ and $\sigma$ are  the  mean  and  standard  deviation  computed  across  the image.
\end{itemize}

In our case, we consider a VGG-19 architecture. The references of the chosen layers are written in the appendix. 
Relying on statistics rather than on a spatial criterion makes the score invariant by translation of different components of the reference image.
We used both low level or  relatively high-level features of VGG-19. As shown in \cite{zeiler2014visualizing}, low level layers catch very simple patterns like edges, specific shades of color, transitions, etc. The higher a layer is in the network, the more abstract, large and complex the pattern it reveals is. Some of these patterns are very specific, leading to some neurons mostly activated by faces for example. 

\subsubsection{Latent vector penalty}

The PGAN model normalizes the latent vector before feeding it to the generator.
\ccc{Without any further penalty on the latent vector, \TIPanswer{the magnitude of} $z$ could shrink \TIPanswer{or explode}. To avoid that,} we add a criterion constraining the norm of the latent vector to be one:
\begin{equation}
    \mathcal{C}_{\nu} = \left( \frac{1}{N}\| z \|^2 -1 \right)^2,
\end{equation}
\cc{where $N$ is the size of $z$.}

\subsubsection{Realism penalty}
Finally, if $\mathcal{C}_S$ forces the generated image to show some statistical similarity with the reference, nothing compels the generator to produce a realistic output. That is why we complete our score with a realism criterion~$\mathcal{C}_R$, that exploits the trained discriminator as well:
\begin{equation}
    \mathcal{C}_R = - D (G(z)).
\end{equation}

\subsubsection{Final penalty}

Our criterion becomes:
\begin{equation}
    \mathcal{C} = \lambda_{S} \mathcal{C}_S + \lambda_{\nu} \mathcal{C}_{\nu} + \lambda_{R}\mathcal{C}_{R} + \lambda_{L} \mathcal{C}_L,
\end{equation}

where $ \lambda_{S},  \lambda_{\nu},  \lambda_{L}, \lambda_R$ are positive scalars. \TIPanswer{We scaled these scalars to ensure that the variations of each term are of the same order of magnitude.}
Our goal is to find the optimal solution
\begin{equation}
	z^* = \argmin_z \mathcal{C}(z,I). \label{tutu}
\end{equation}

\subsubsection{Inspiration with a class-conditioned generator}

When a generator \ccc{is} trained using a class input in addition to $z$ (AC-GAN \cite{Odena2017conditionalGAN}), the class conditioning part of the latent vector is supposed to take discrete values. 
To overcome this problem, we tried two strategies:
\begin{enumerate}
    \item Performing the search normally to find a continuous-valued vector consisting of the concatenation of $z$ and input class vector $c$. Unless specified otherwise, we always adopt this strategy in our experiments.  
    \item Holding the user provided input class fixed, and performing the gradient descent only on the latent part $z$. We denote this setting as the ``conditioned" one.
\end{enumerate}

\subsection{Algorithm selection}

\ccc{Beyond the classical Random Search (RS) baseline, that consists of repetitively sampling random $z$ given a budget of $N$ iterations to lead to a retrieved $G(z)$, we consider gradient-based and evolutionary approaches. }  

\subsubsection{Gradient-based retrieval given a target}

We experiment with classical optimization algorithms such as the Adam method \cite{adam}, and LBFGS \cite{lbfgs}, a well know quasi-Newton algorithm approximating second order information with successive values of the gradient. The choice of Adam as an optimizer may seem odd since our problem is not stochastic. However, the function we consider in this problem shows local variation similar to noise. Using Adam is a way to smooth the optimization process.
However, LBFGS has a better convergence rate than Adam: $O(1/t)$ with $t$ the number of iterations. Therefore we expect it to show better performance. We also include Nesterov's momentum, with various learning rates and momentum 0.9, advocated for in \cite{fairlatent}.

\subsubsection{Gradient-free optimization}\label{whynogradient}

In addition to the optimizers tested above, we tried out several gradient-free optimizers in our image retrieval task. Indeed, for solving a non-convex problem, gradient-free optimizers could outperform more standard methods. Besides, at equal number of steps, they are computationally faster than gradient-based methods. 
After several tests, we selected \cc{the} 
three \cc{best performing} gradient-free optimizers from the Nevergrad optimization \cite{nevergrad} library:
\begin{itemize}
	\item Two variants of the Differential Evolution method \cite{de} (DE): 2-points DE (2PDE) which replaces the classical pointwise crossover of DE by a two-point crossover \cite{holland} as advocated for in \cite{nevergrad}, and DDE (Discrete-DE) which uses in DE the crossover rate $1/d$ with $d$ the dimension of $z$, which is classical in discrete optimization. 
    \item The discrete $(1+1)$-evolution strategy \cite{discrete} (DOPO). {DOPO is a special case of the $(\mu/\mu+\lambda)$-optimization described in Algorithm \ref{fcalg}, with $\mu=\lambda=1$. $1 + \lambda$ candidate latent vectors are generated among which the $\mu$ best are selected according to the loss criterion defined above. From this selection, a new value of the optimal vector $\hat{z}$ is inferred. The process is repeated until $G(\hat{z})$ is satisfying.}
\end{itemize}

\begin{algorithm}
\otcxx{
\begin{algorithmic}
    \STATE $\forall 1\leq j \leq \mu, \hat{z_j} := (0,\dots,0) \in \mathbb{R}^d$.
	\WHILE{Computational budget not elapsed}
	\FOR {$i\in \{1,\dots,\lambda\}$}
		\STATE $z_i := \hat z_j$ for $j\leq \mu$ and $\mu$ divides $i-j$
		\ENDFOR
			\FOR {$i\in \{\mu+1,\dots,\mu+\lambda\}$}
		\FOR{$ p \in \{1,\dots,d \}$}
    		\STATE With probability $1/d$ replace $z_i(p)$ with an independent standard normal value.
		\ENDFOR
		\STATE repeat if no variable $z_i(p)$ was modified.
	\ENDFOR
	\STATE $z_{\lambda +1 } := \hat{z}$
	\STATE Select the $\mu$ best images $G(z_{\alpha_1}), \dots G(z_{\alpha_\mu})$ among $G(z_{1}), \dots G(z_{\lambda + \mu})$
	\STATE $\forall 1\leq j\leq \mu, \hat{z}_j := z_{\alpha_j}$ %\frac{1}{\mu} \sum_{i=1}^\mu z_{\alpha_i}$
	\ENDWHILE
	\RETURN $\hat{z}_1$
\end{algorithmic}}
	\caption{\label{fcalg}$(\mu/\mu+\lambda)$-optimization, with a latent space of dimension $d$.}
\end{algorithm}

Numerical scores usually do not perform as well as human judgment: with human input we could hope to reach more visually convincing results. This is why we \ccc{introduce} a last optimizer, HEVOL (Human in the EVOlution Loop), where a human user chooses manually at each iteration the $\mu$ best out of $1 +\lambda$ possibilities.%Typically for face generation, we took $\mu= 5$ and $\lambda=27.$
\omitme{,or $\mu=1$ and $\lambda=15$ (fashion generation)}
A target image is not even necessary with this method: the user could also reconstruct an image they have in mind, or just provide preferences \ccc{between images}. This is convenient for applications such as facial composites or fashion generation (Section \ref{fc} and Section \ref{fir}).

%---------------------------------
\section{Experimental setting}
%---------------------------------

\subsection{Datasets}
We trained generators on four publicly available datasets, two small ones and two large ones:
\begin{itemize}
    \item The Describable Textures Dataset (DTD): a dataset of texture patterns \cite{cimpoi14describing} consisting of 5640 images labeled into 47 different categories. 
    \item The RTW dataset described in \cite{sbai2018design}, consisting of 4157 images of plain clothes in front of a white background, labeled with 7 shapes and 7 texture categories.
    \item The Celeba-HQ dataset introduced by Karras et al. \cite{karras2017progressive}, consisting of 30,000 images of celebrity faces. We do not consider the available labels for this data-base.
    \item The FashionGen dataset \cite{fashionGen2018}  containing 293,008 labelled fashion images. Each sample contains one item, worn by a model in the case of clothes, in front of a white background. We trained our model on the clothing sub-dataset with about 200,000 items.  
\end{itemize}

\subsection{Target images: reconstruction, semi-specified, misspecified cases}

We distinguish the reconstruction case (the target image is randomly generated by the generator trained on the dataset), the semi-specified case (the target belongs to the training dataset), and the misspecified one (the target comes from another source, e.g. Wikipedia, \brtwo{random face pictures} \cite{karras2018style} or DeepFashion \cite{liuLQWTcvpr16DeepFashion}). \omitme{Table \ref{misscases}} Our results show that scores degrade from specified to misspecified, which is close to semi-specified.

\subsection{Hyper-parameters}
We used default values of progressive growing of GANs. The dimension of $z$ is 512. Most models have been trained until a 256$\times$256 resolution though we ran generations up to 512$\times$512 on the RTW and CelebA-HQ datasets. \cc{To obtain faster results, we down-scaled the inputs as described in the Appendix.}
For all gradient-based optimizers, we worked with a batch size of~1. For Adam, we kept the hyper-parameters used for the training of progressive growing ($(\beta_1, \beta_2) = (0, 0.99)$) and tested various learning rates. For Nesterov momentum, we use momentum 0.9 and various learning rates. In all cases we started with a base gradient step equal to one. We perform gradient decay twice, dividing the gradient step by 10 at one third of the iteration budget, and then again at two thirds. 
To obtain better results, we reset the model's state to the one associated with the smallest loss found so far before each decay of the gradient step.
We performed a grid search to select the best gradient step for Adam, Nesterov, and LBFGS. Unless specified otherwise, all optimizations are launched with 1000 iterations.
All gradient-free optimization methods have been tested with their default parameters in \cite{nevergrad}. \omitme{which matches standard values in the field An exception is DDE, which uses DE with the crossover-rate equal to $1/d$ with $d$ the dimension.}

%----------------------------------------------------------------------

\section{Results}

\subsection{Criterion balance}

\label{sec:details_insp}

\begin{table}
	\center
\begin{tabular}{cccc}
\hline
             & L2 & L2 + VGG & VGG \\
             \hline
$\lambda_L$ & 50 & 50 &  0 \\                
$\lambda_R$  & 0 & 0.1 & 0.1 \\
$\lambda_S$ &0 & 1 & 1 \\
\hline
\end{tabular}
\vspace{0.5ex}
	\caption{Default parameters for each criterion.}
	\label{criteriatab} 
\end{table}

\begin{figure}
\center
\begin{tabular}{c}
     \includegraphics[width=1\linewidth]{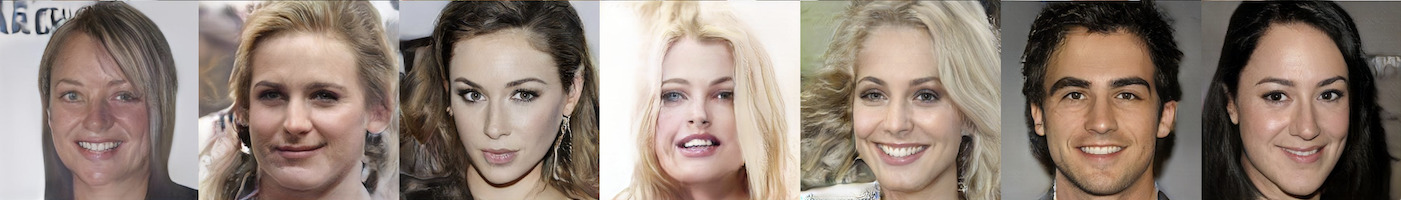}
\end{tabular}
\caption{Random samples sorted by increasing value of $D$. We observe that most images containing artifacts have a small $D$, highlighting the importance of the discriminator part in our criterion. \label{fig:realism}}
\end{figure}

	We use various combinations of similarity measures, namely the simple $\ell_2$ loss $\mathcal{C}_L$, the  feature similarity $\mathcal{C}_{S}$ loss, the discriminator realism criterion $\mathcal{C}_R$. They are presented in Table~\ref{criteriatab}.
 We found in our trials that $\lambda_{\nu} =1$ was an acceptable value for all models. The adequate value of $\lambda_R$ depends on the training configuration of the GAN and more precisely on the discriminator's output. If $D$ is ``too strong'' and returns very high scores in absolute value, then $\lambda_R$ must be reduced accordingly. For Celeba-HQ and FashionGen-clothing, we selected $\lambda_R \approx 0.1$, as it leads to slighly less artifacts (see Figure \ref{fig:realism}). \omitme{However, as shown in Figure \ref{fig:realsim2}, a compromise must be found between similarity and desired realism.}

\def\popo{
\subsection{Recovering latent variables}
TODO OT zdist
Only LBFGS, and mainly with L2 criterion, successfully recovers latent variables TODO}

\paragraph{Human study}

\begin{table}[htb]
	\center
\begin{tabular}{|c|cccc|}
\cline{2-5}
\multicolumn{1}{c|}{} & L2 & L2+VGG & VGG & VGG no R \\  
\hline
cond LBFGS & 0.8 & 13.6 & 5.0 &  26.1 \\       
cond DOPO  & 2.3 & 12.7& 10.4& {\bf 29.2} \\
\hline
\end{tabular}\\
\vspace{0.5ex}
a) Criterion ablation study on DTD. \\ 
~\\
\begin{tabular}{|c|ccc|}
\cline{2-4}
\multicolumn{1}{c|}{} & L2 & L2+VGG & VGG\\  
\hline
LBFGS & 7.8 & 13.5 & 14.7 \\ 
2PDE & 8.5 & {\bf 35.2} & 20.3 \\
\hline
\end{tabular}\\
\vspace{0.5ex}
b) Criterion ablation study on the FashionGen dataset\\

	\caption{Selection of the best criterion setting for DTD (semi-specified) and (misspecified) FashionGen image retrieval: human scores (\% of retrieved images judged most similar to the target).}
	\label{res1}
\end{table}

To validate the best criterion and algorithm for each case we conducted a human study. For each of 100 given targets, we asked five different participants to pick the most similar image between images generated with different criteria, or optimizers. 
For each dataset, we first identified the two best performing algorithms by visual inspection and conducted the study to identify the best criterion between the different possibilities of Table \ref{criteriatab}. Table \ref{res1} presents the results on the DTD (semi-specified setting) and FashionGen dataset (misspecified setting). The best identified criterion for DTD is the VGG without Realism penalty, and on FashionGen, the L2+VGG setting.

\begin{figure}
\center
\setlength\tabcolsep{1.5pt}
\begin{tabular}{lcccc}
\textbf{Target} & \textbf{2PDE}& \textbf{DOPO} & \textbf{Adam} & \textbf{LBFGS} \\
\includegraphics[width=0.18\linewidth]{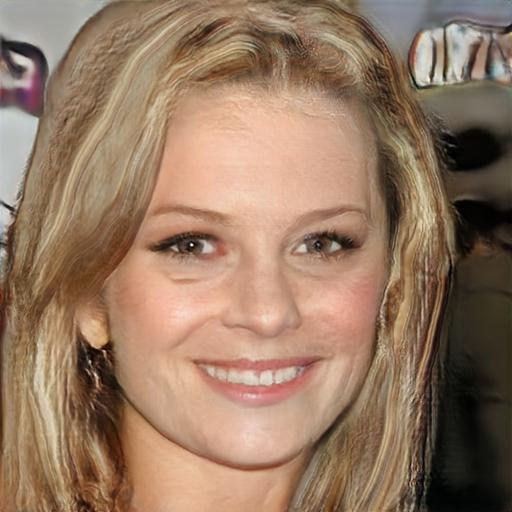}~~ &
\includegraphics[width=0.18\linewidth]{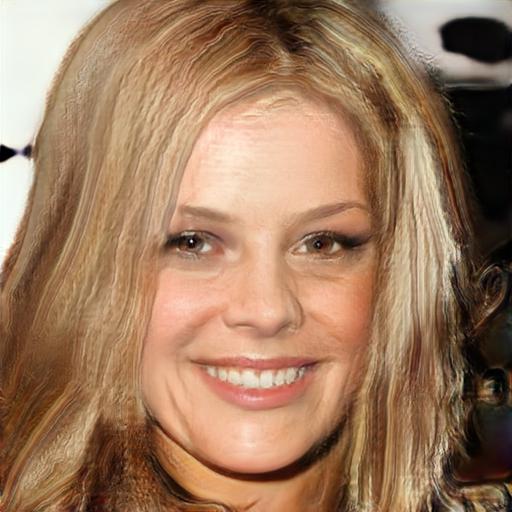} &
\includegraphics[width=0.18\linewidth]{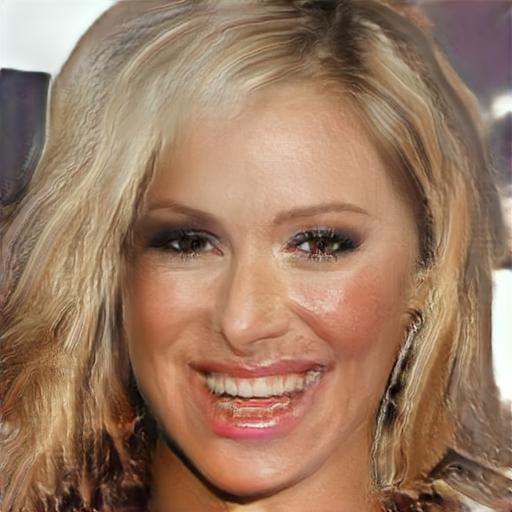}&
\includegraphics[width=0.18\linewidth]{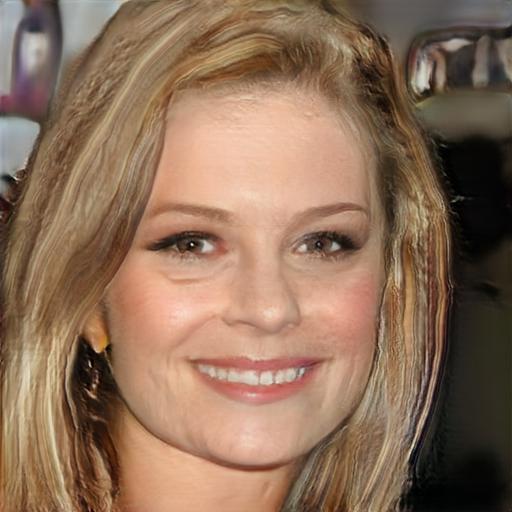}&
\includegraphics[width=0.18\linewidth]{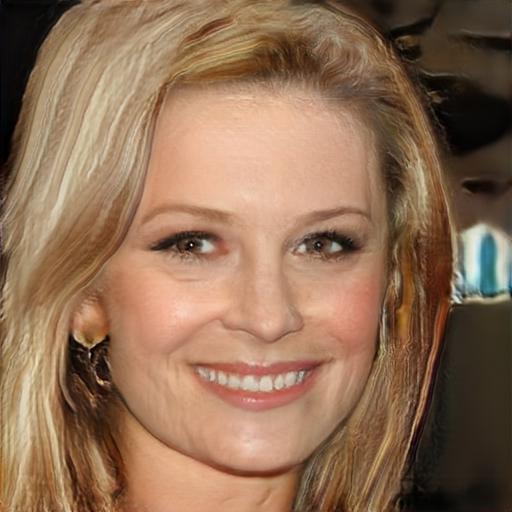}\\
\includegraphics[width=0.18\linewidth]{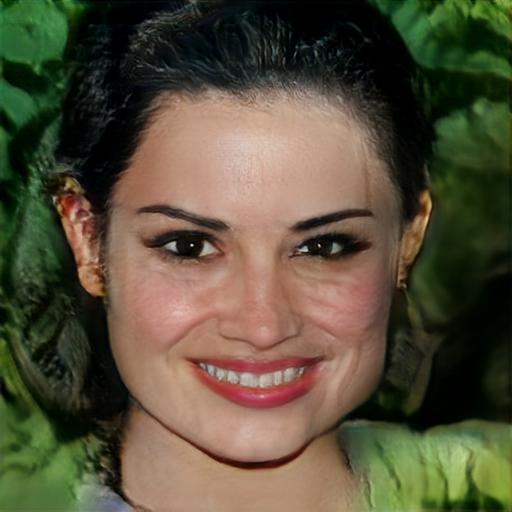}~~ &
\includegraphics[width=0.18\linewidth]{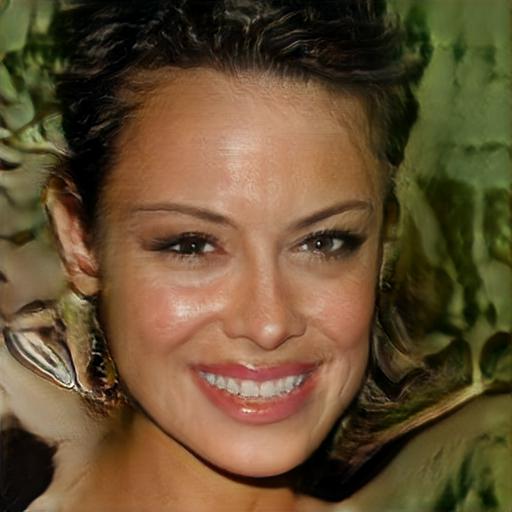} &
\includegraphics[width=0.18\linewidth]{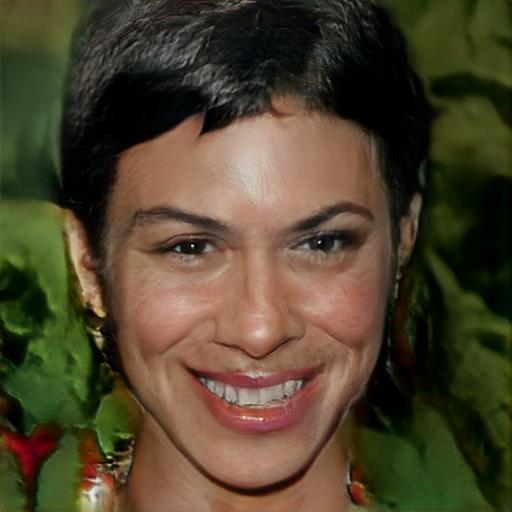}&
\includegraphics[width=0.18\linewidth]{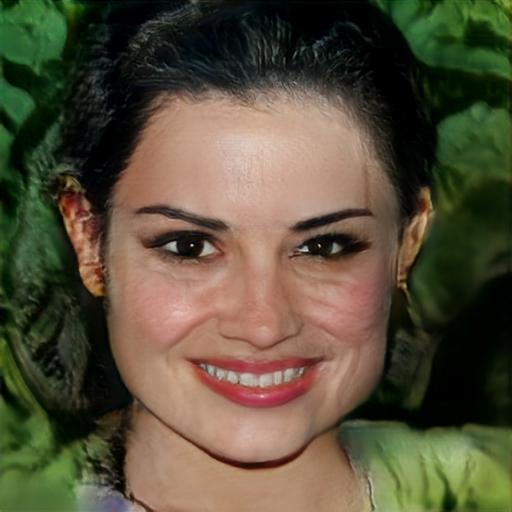}&
\includegraphics[width=0.18\linewidth]{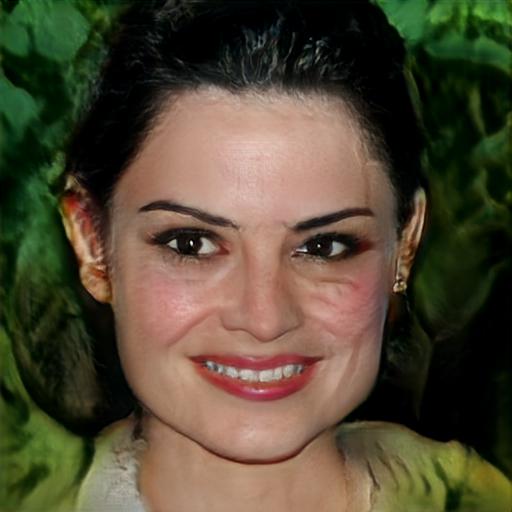}\\
\end{tabular}
\caption{\label{fig:reconstruction1}\brtwo{Reconstruction of images generated with a GAN trained on CelebA-HQ} using different optimizers (3000 iterations, VGG loss). We observe superior performance of gradient-based optimizers\otc{, in particular for LBFGS which approximates a second order method\br{.}\omitme{ - t} To the best of our knowledge, this is the first application of LBFGS for searching the latent space, and\br{, numerically,} it clearly outperforms\omitme{, numerically,} other methods (see also Figure~\ref{reconst}}).}
\end{figure}

\begin{figure}
\center

\includegraphics[width=0.16\linewidth,height=0.16\linewidth]{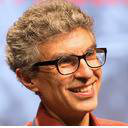}
\includegraphics[width=0.16\linewidth,height=0.16\linewidth]{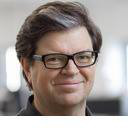}
\includegraphics[width=0.16\linewidth,height=0.16\linewidth]{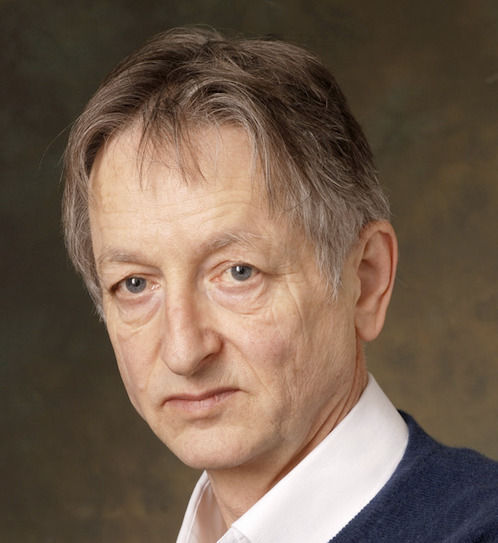}
\includegraphics[width=0.16\linewidth,height=0.16\linewidth]{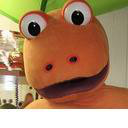}
\includegraphics[width=0.16\linewidth,height=0.16\linewidth]{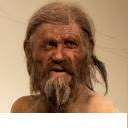}\\
\includegraphics[width=0.16\linewidth,height=0.16\linewidth]{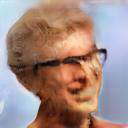}
\includegraphics[width=0.16\linewidth,height=0.16\linewidth]{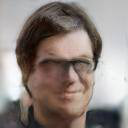}
\includegraphics[width=0.16\linewidth,height=0.16\linewidth]{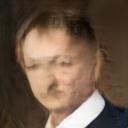}
\includegraphics[width=0.16\linewidth,height=0.16\linewidth]{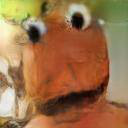}
\includegraphics[width=0.16\linewidth,height=0.16\linewidth]{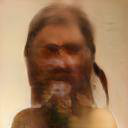}\\

\caption{Targets (first line) and images retrieved (second line) using an L2 similarity criterion and the LBFGS algorithm with 5000 iterations with a model trained on CelebaHQ. \label{celebaL2} \otc{Even characters very far from the original dataset (such as \"Otzi) or not even human (Casimir, the gentle dinosaur) can be reconstructed if we have no other penalty than L2. \omitme{- which leads to an expensive copypasting}}}
\end{figure}

\subsection{Impact of the VGG and L2 terms}

\begin{table}[]
    \centering
    {\small 
    \begin{tabular}{cccccc}
    \hline
    Algorithm & Adam  & RS & DOPO  & DDE & LBFGS  \\
        \hline
    reconstruction     & 17.0 & 52.2 & 37.9 & 34.8 & 14.5 \\  
    semi-specified     & 23.5 & 63.0 & 50.6 & 46.5 & 20.6 \\
    misspecified     & 25.5 & 64.0 & 49.5 & 46.5 & 19.6 \\
    \hline
    \end{tabular}}
    \vspace{0.5ex}
    \caption{\label{misscases}  Best score (defined criterion L2 + VGG loss) obtained by various algorithms for recovering faces with a generator trained on Celeba.  Semi- or miss- specified cases are similar, both harder than the reconstruction case. Visual inspection confirms these results: the reconstruction configuration leads to more convincing results. } % Budget 1000, average over 10 images. RS:Random search.
\end{table}

As far as reconstruction is concerned, the feature-based loss gives fair results when combined with Adam or LBFGS (see Figure~\ref{fig:reconstruction1}). However, to perfectly reconstruct the reference image, L2 is the best option to faster retrieve the original output. 
For the semi-specified and misspeficied cases, feature-based similarity criteria are necessary to catch semantically meaningful data regardless of the target's background: haircut, hair color, pose, facial expression, etc. Besides, an L2 loss yields blurry results\omitme{(see Figure \ref{fig:l2vsvgg})}. However, it is interesting to notice that with L2 similarity, GANs can output a rough approximation of pictures sometimes very far from their training dataset (see Figure~\ref{celebaL2}).

\subsection{Optimization method}

\begin{figure}[h!]
\center
\setlength\tabcolsep{1.5pt}
\begin{tabular}{lcccc}
\textbf{Target} & \textbf{2PDE}& \textbf{DOPO} & \textbf{Adam} & \textbf{LBFGS} \\
\includegraphics[width=0.18\linewidth]{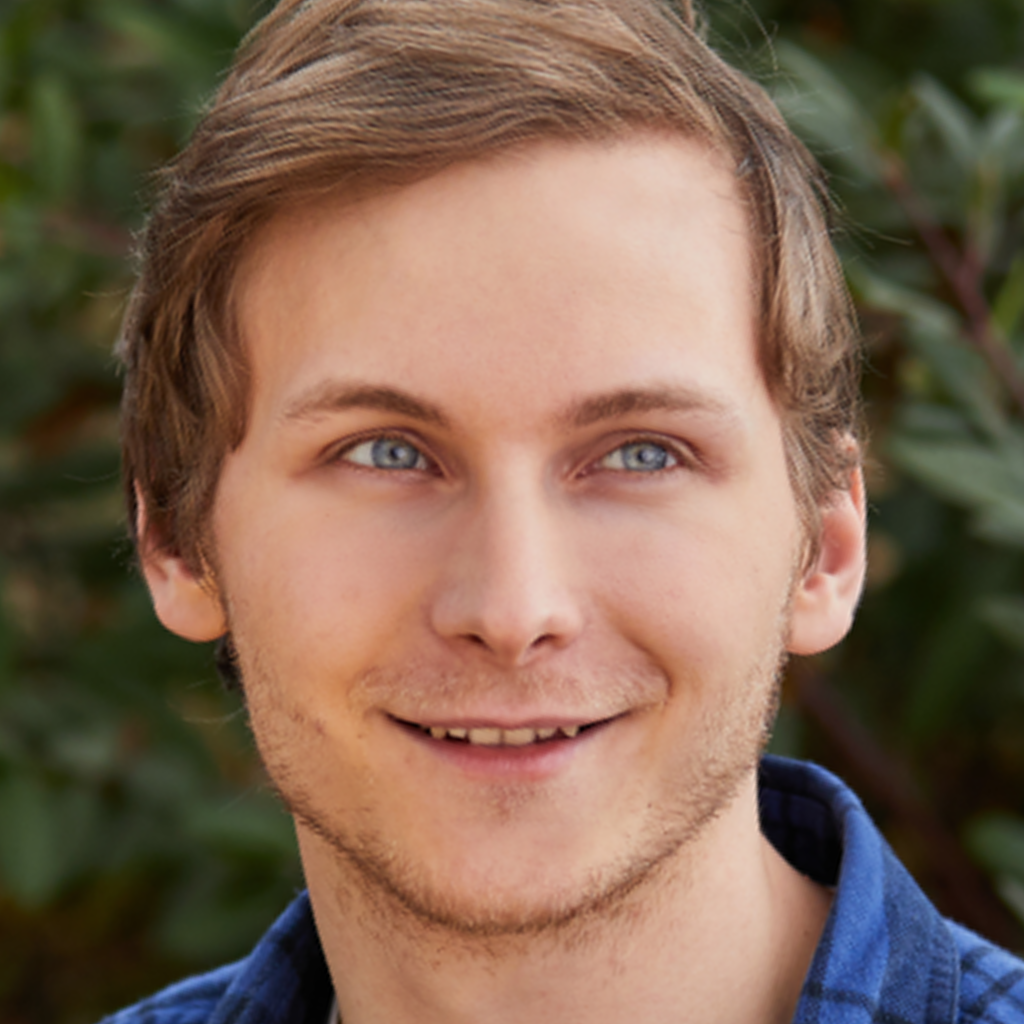}~~ &
\includegraphics[width=0.18\linewidth]{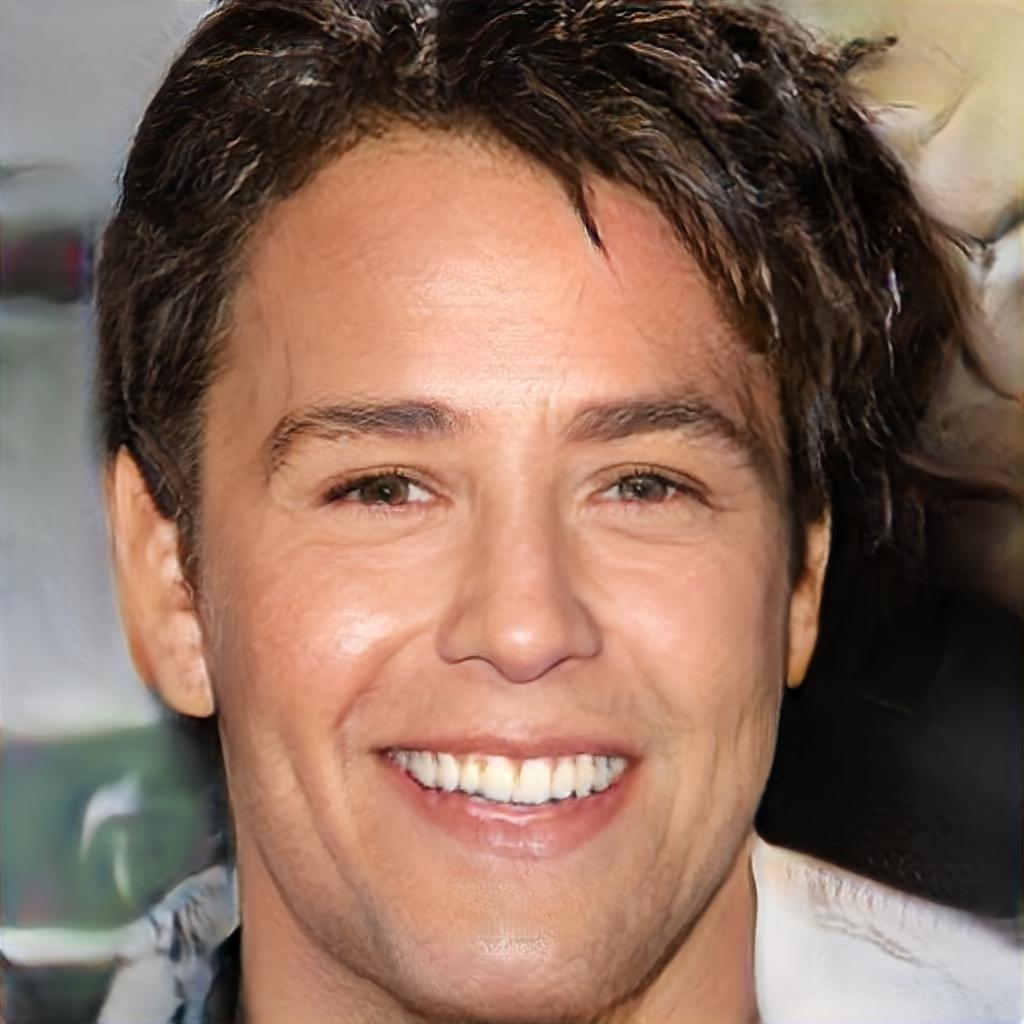} &
\includegraphics[width=0.18\linewidth]{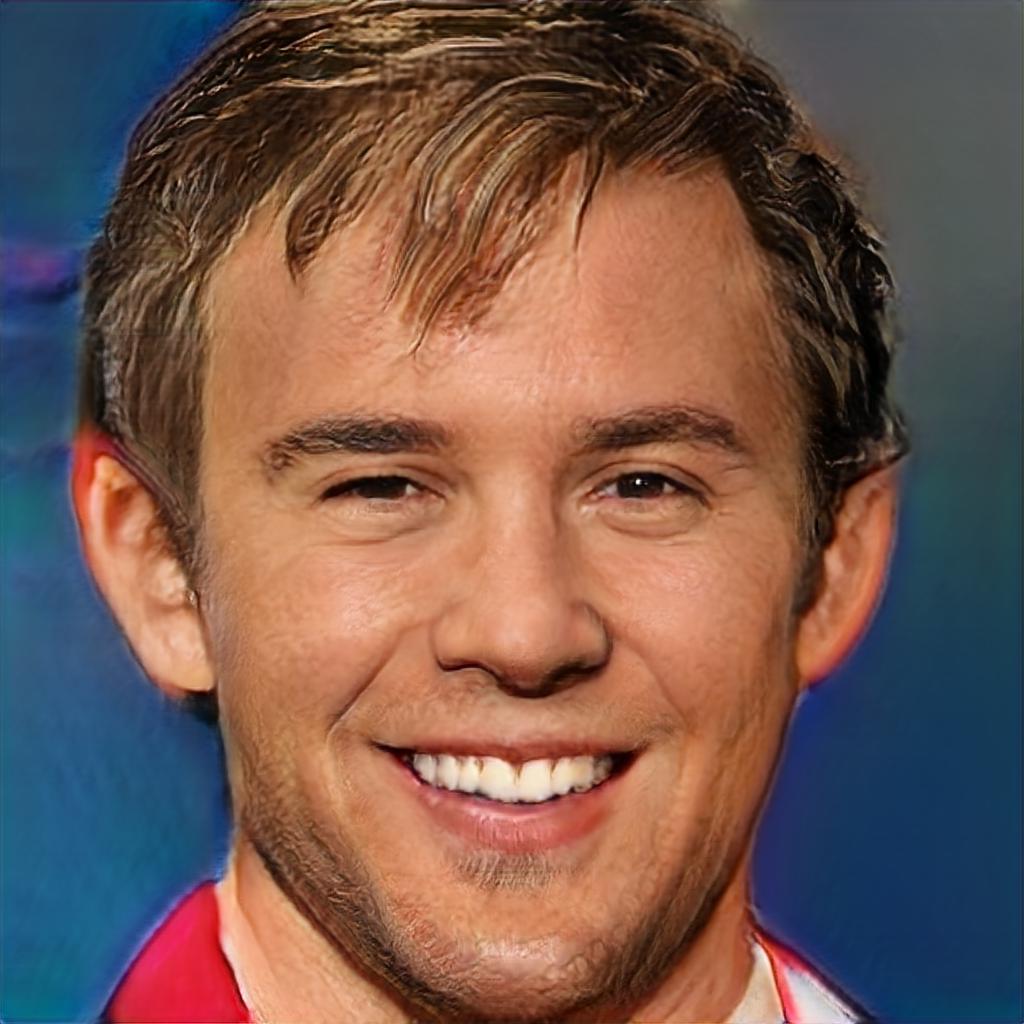}&
\includegraphics[width=0.18\linewidth]{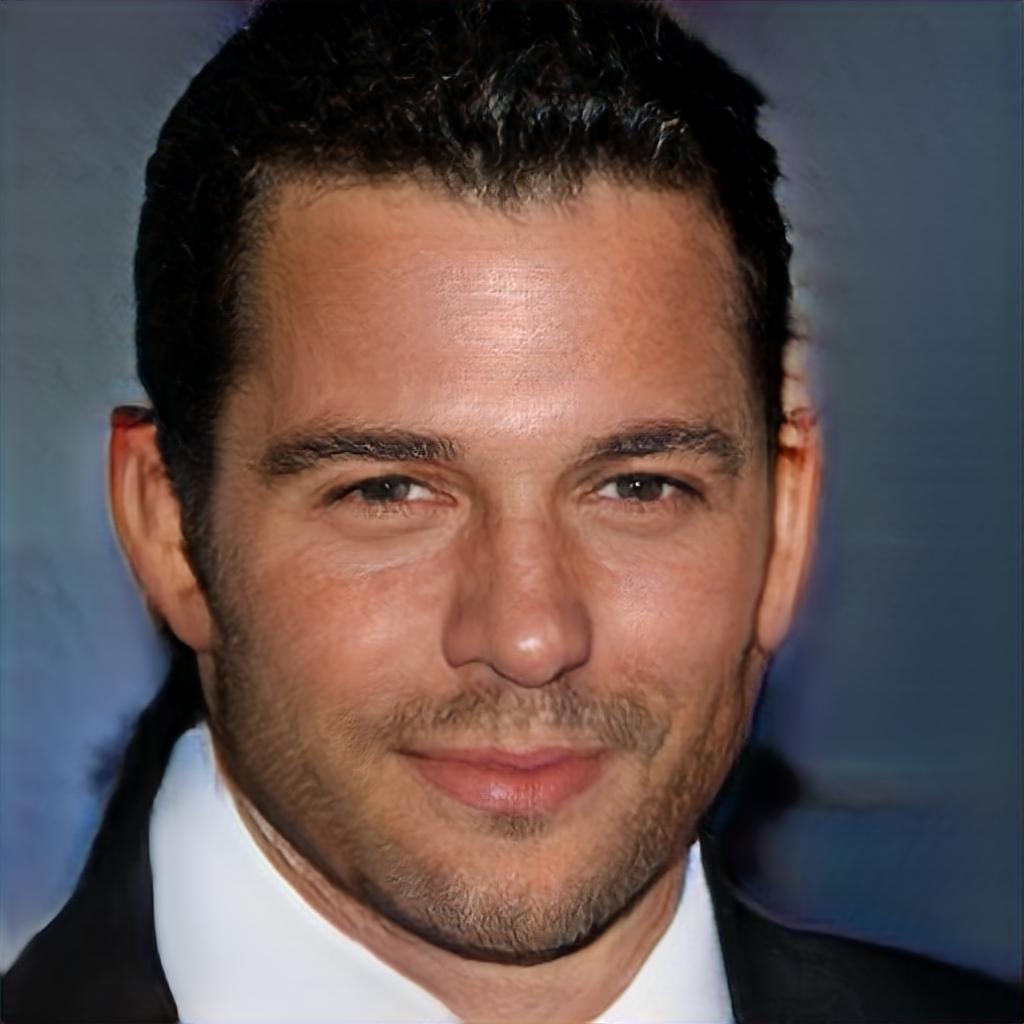}&
\includegraphics[width=0.18\linewidth]{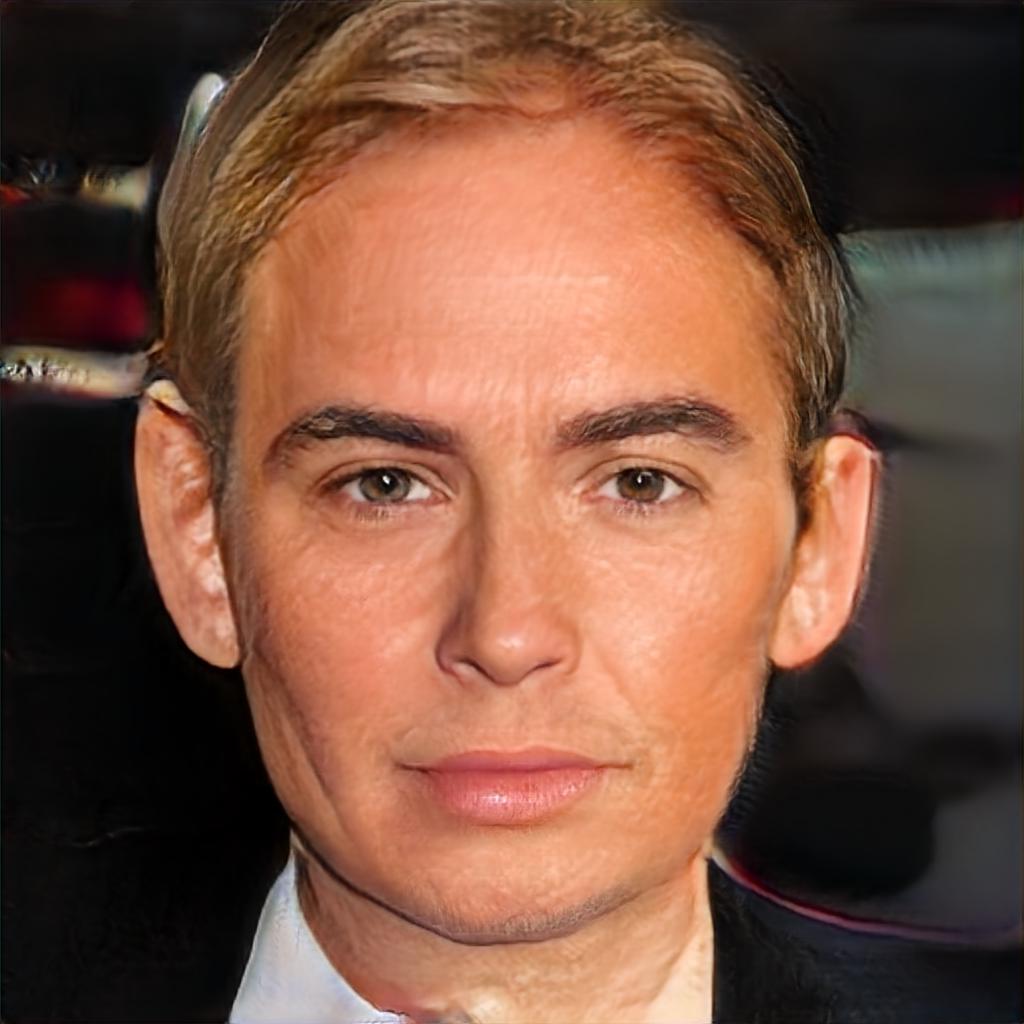}\\
\includegraphics[width=0.18\linewidth]{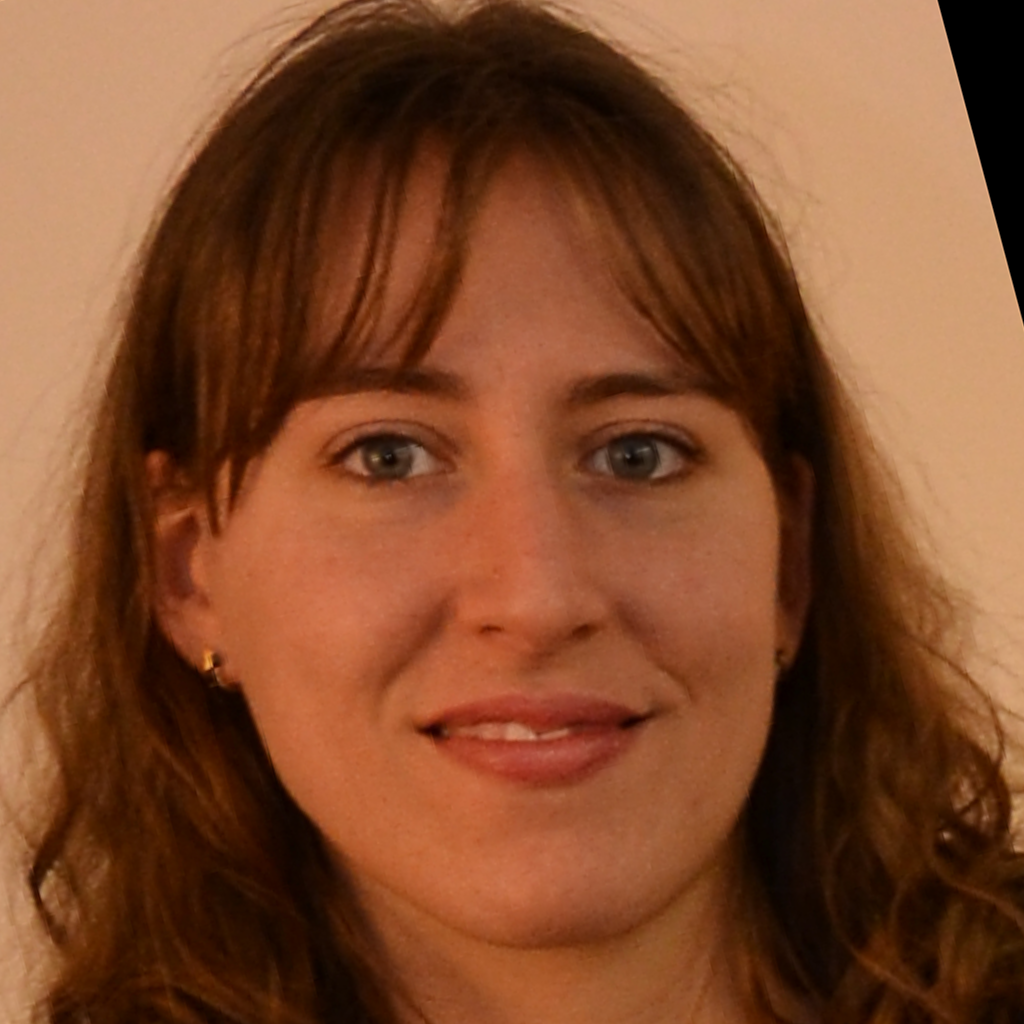}~~ &
\includegraphics[width=0.18\linewidth]{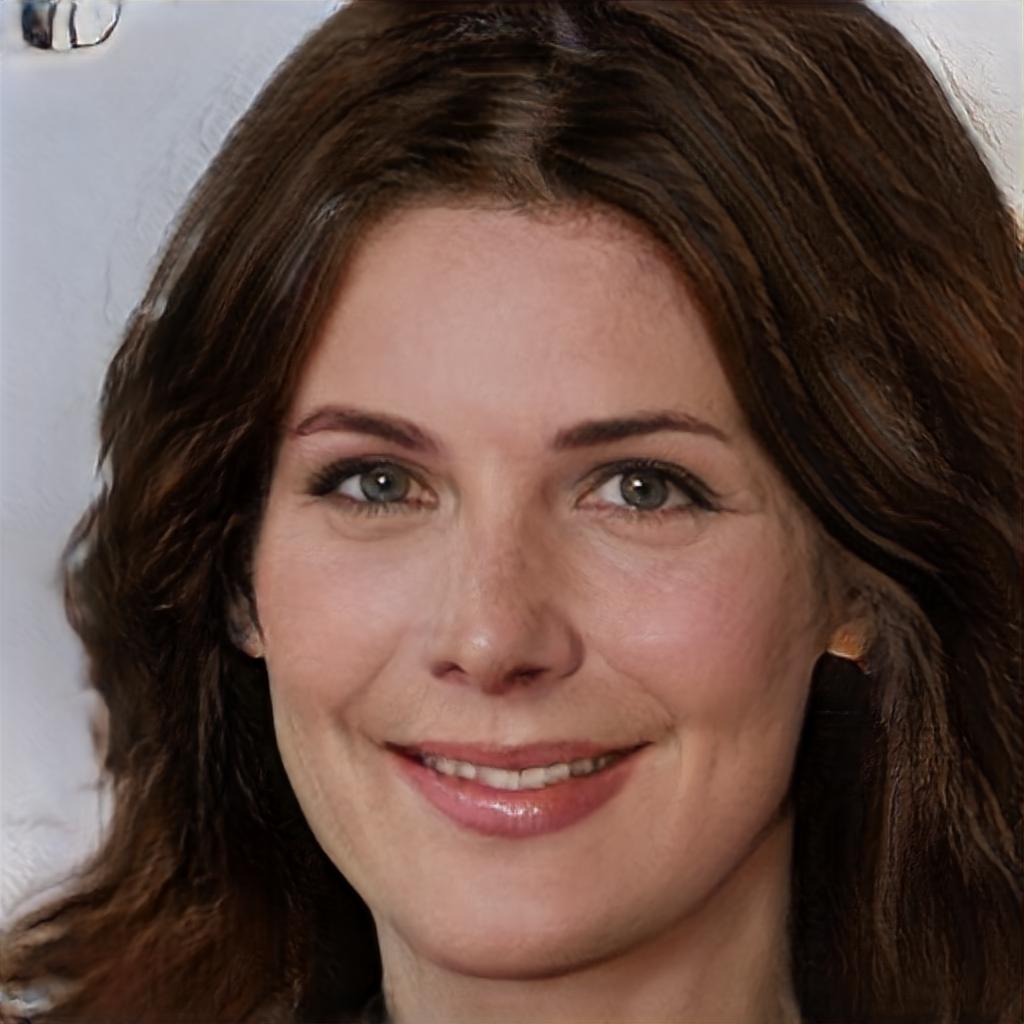} &
\includegraphics[width=0.18\linewidth]{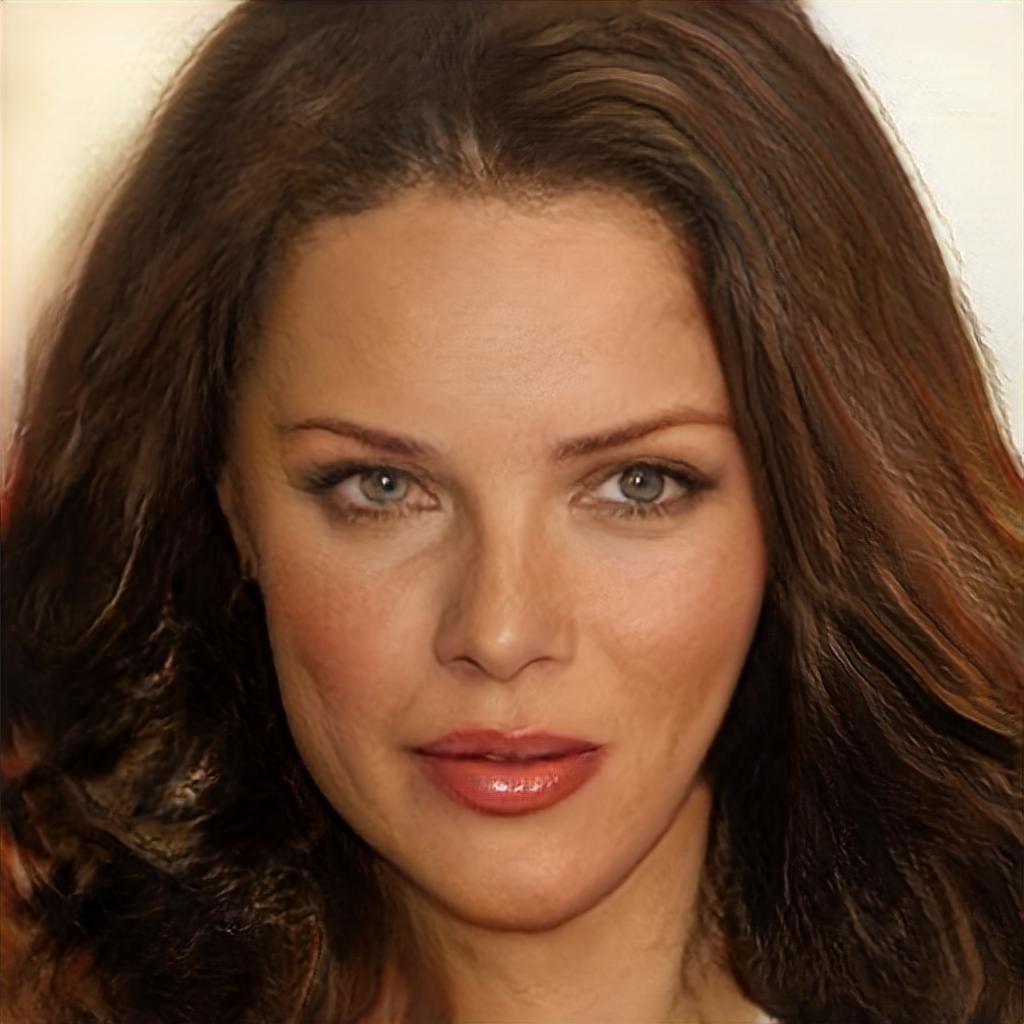}&
\includegraphics[width=0.18\linewidth]{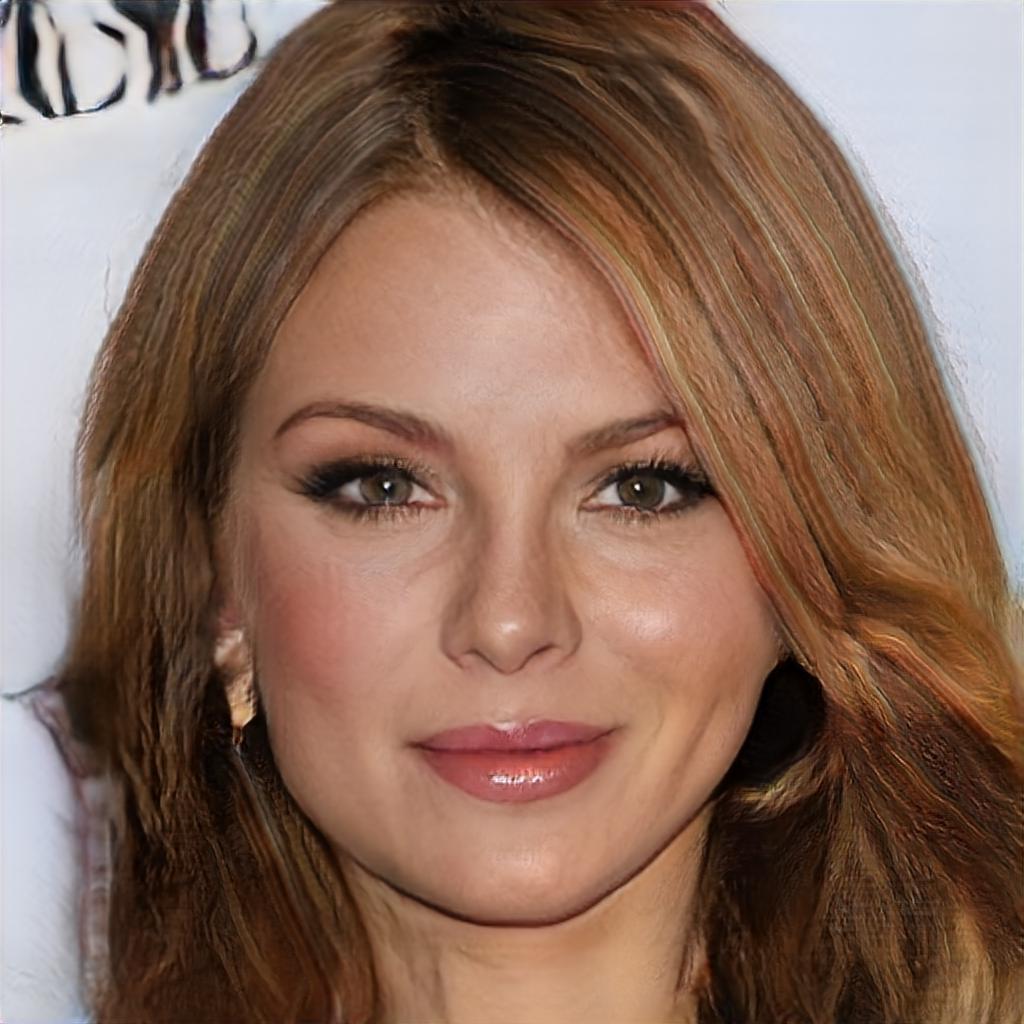}&
\includegraphics[width=0.18\linewidth]{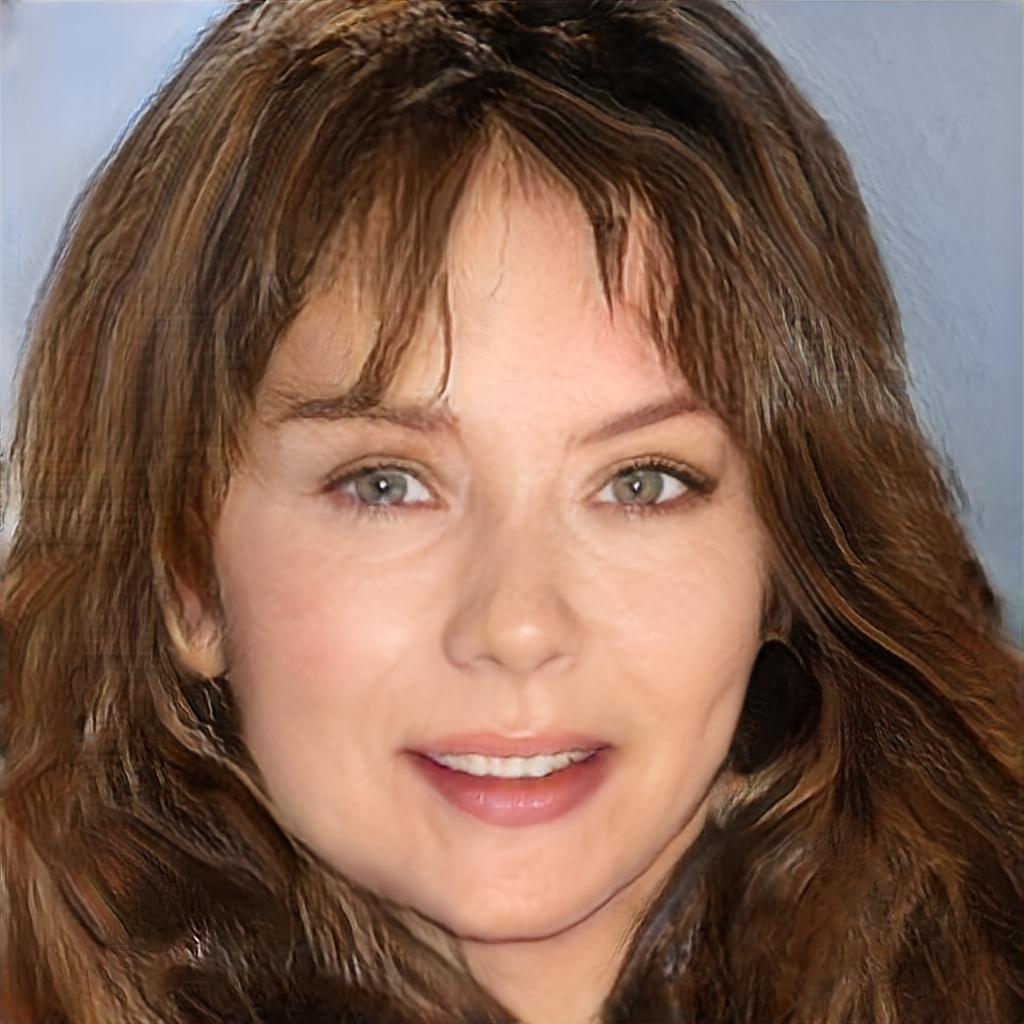}\\
\includegraphics[width=0.18\linewidth]{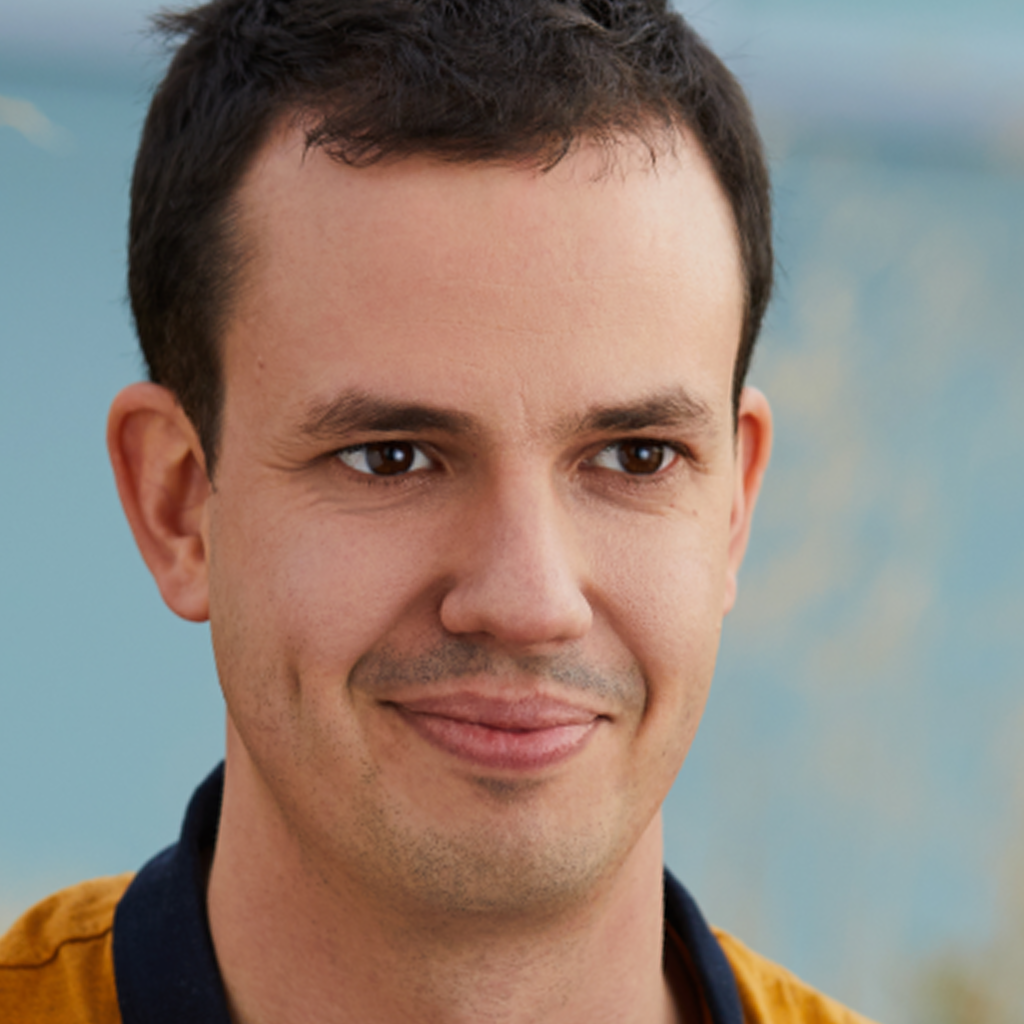}~~ &
\includegraphics[width=0.18\linewidth]{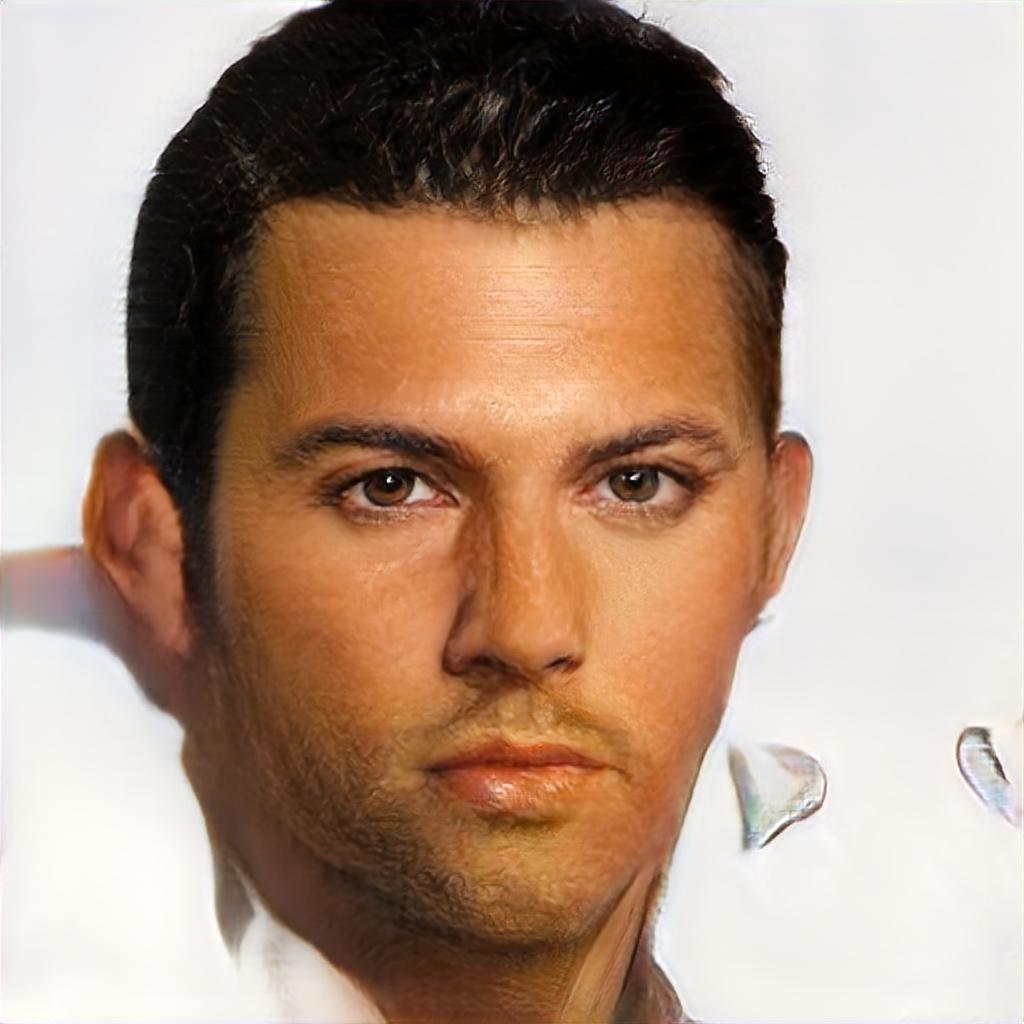} &
\includegraphics[width=0.18\linewidth]{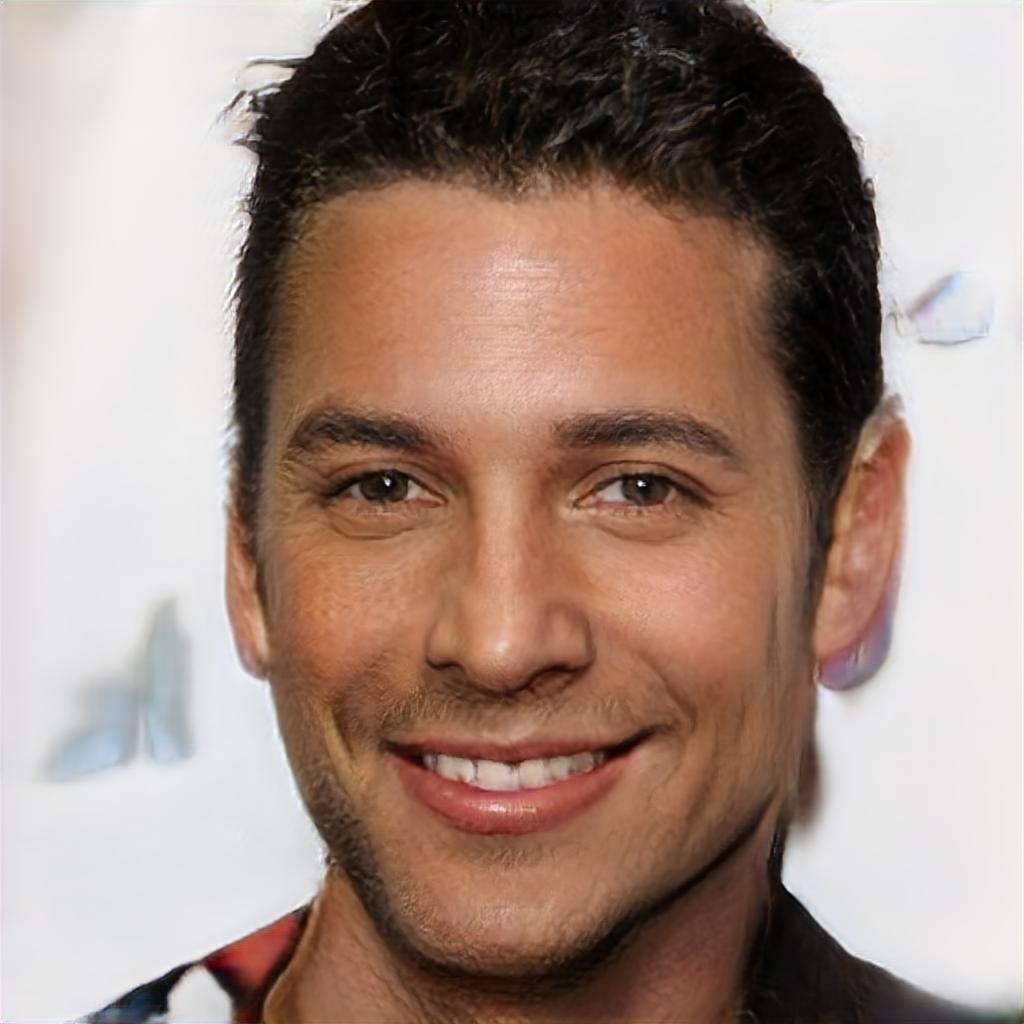}&
\includegraphics[width=0.18\linewidth]{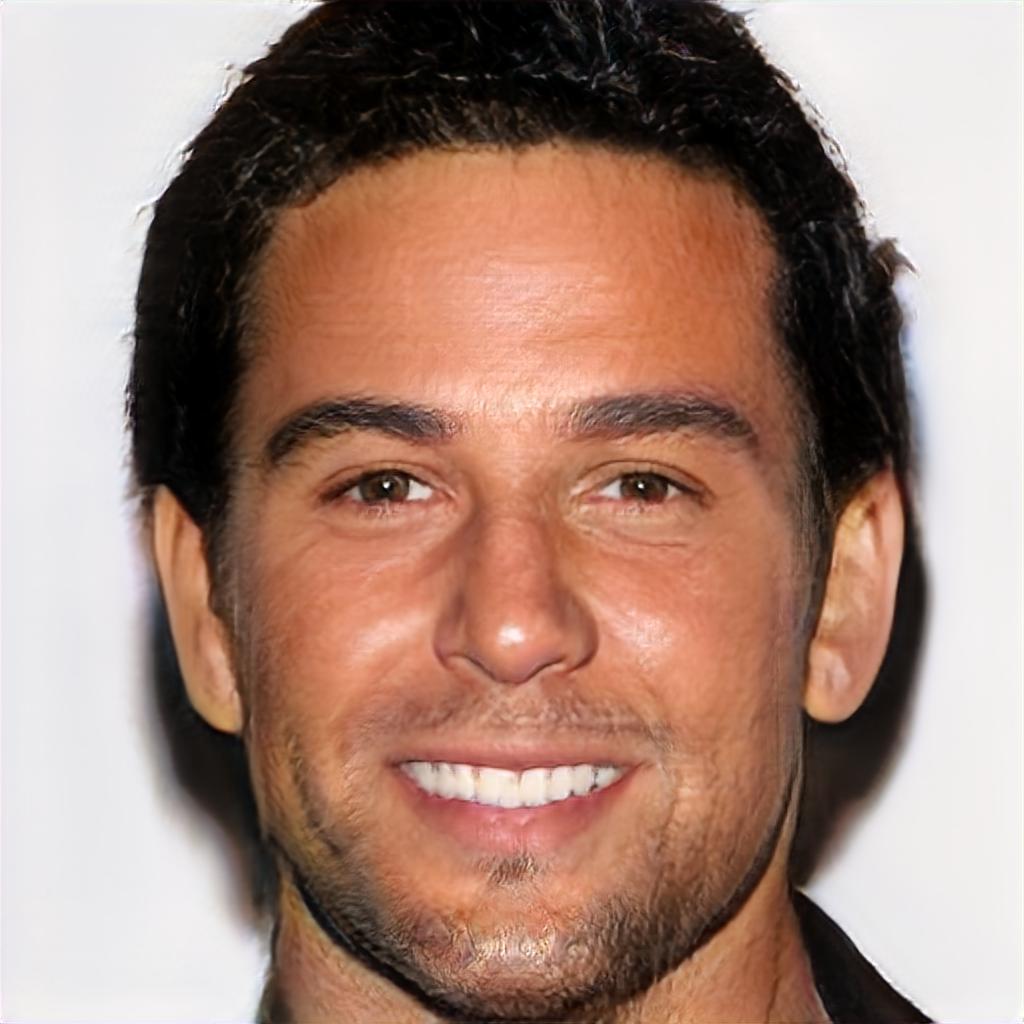}&
\includegraphics[width=0.18\linewidth]{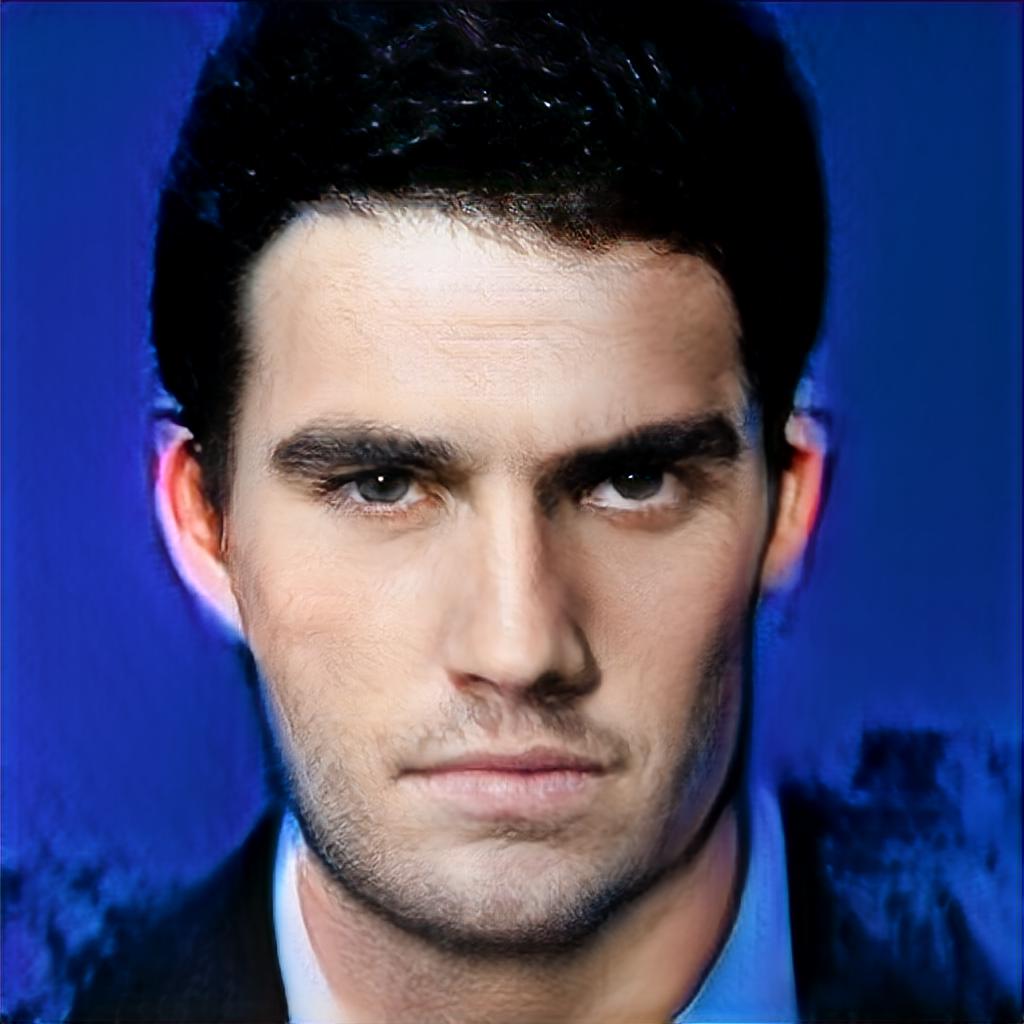}\\
\includegraphics[width=0.18\linewidth]{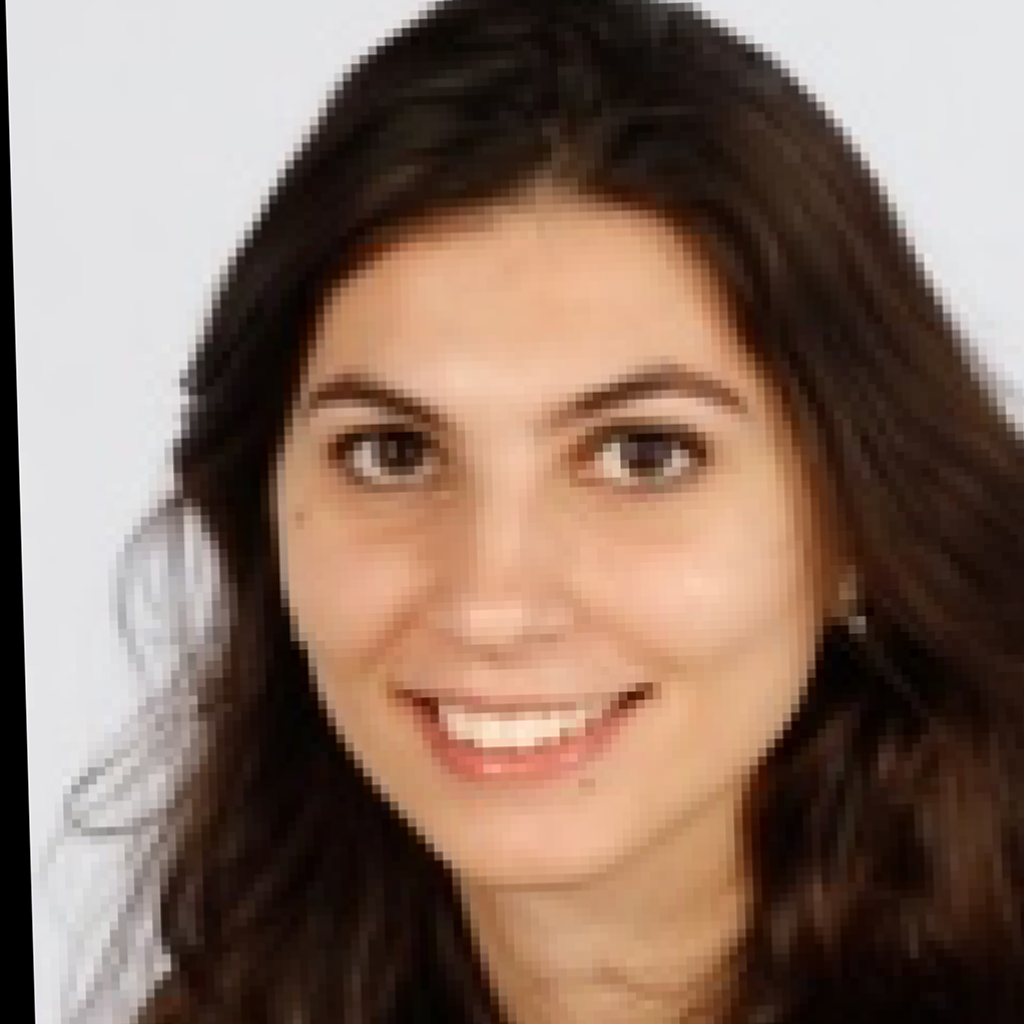}~~ &
\includegraphics[width=0.18\linewidth]{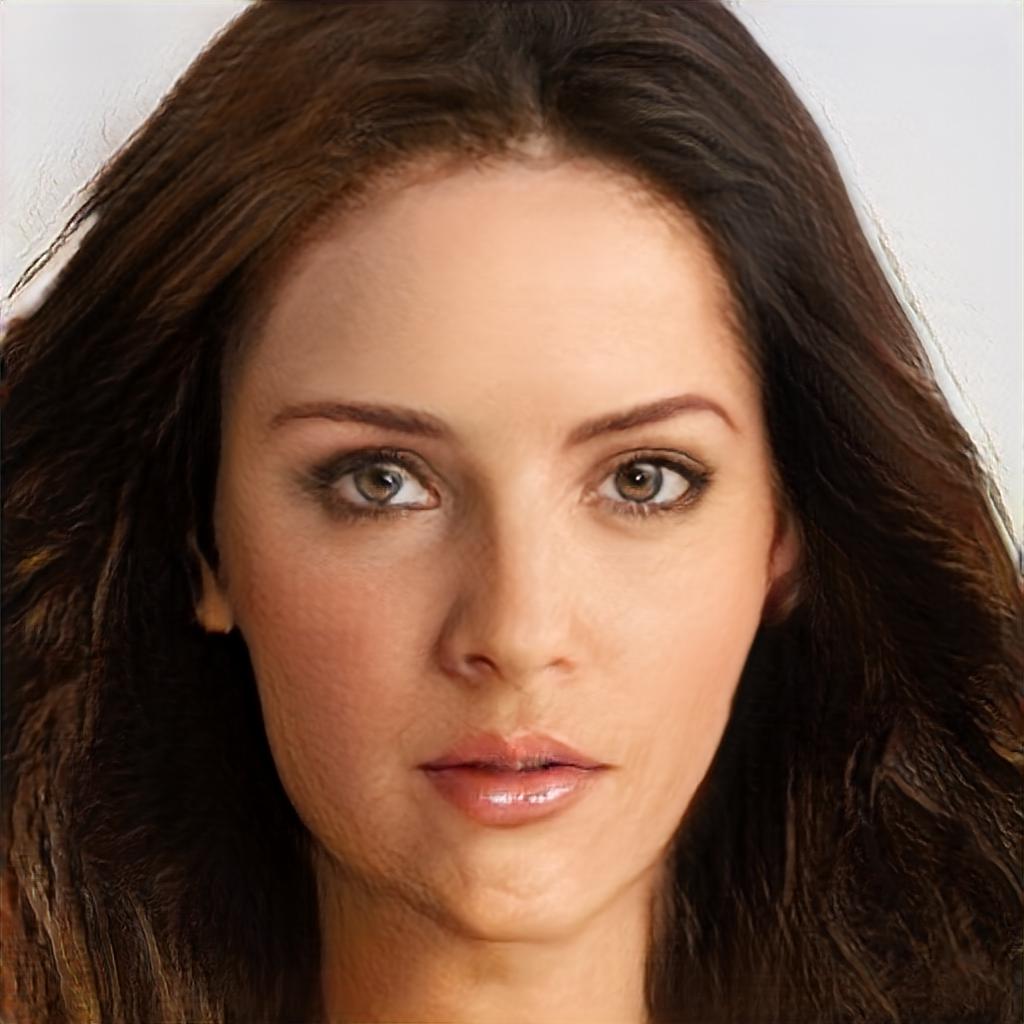} &
\includegraphics[width=0.18\linewidth]{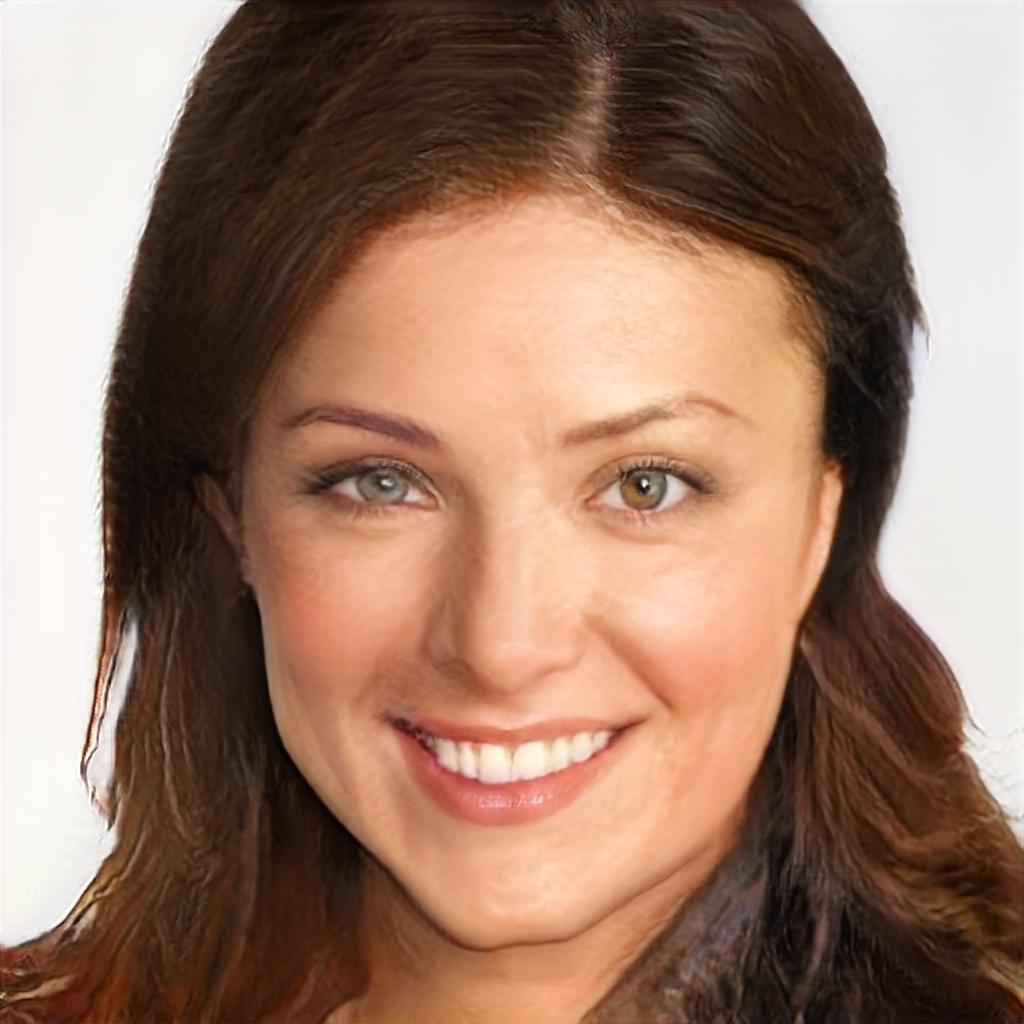}&
\includegraphics[width=0.18\linewidth]{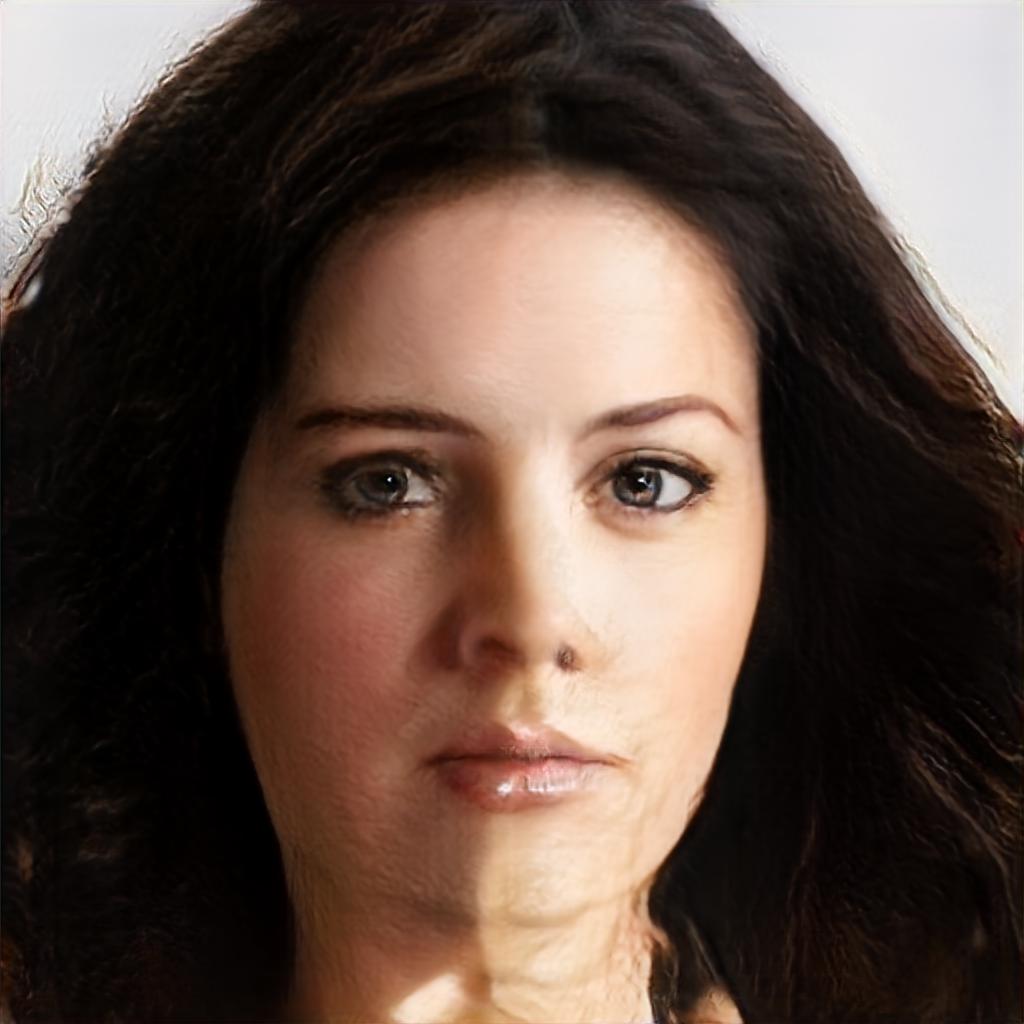}&
\includegraphics[width=0.18\linewidth]{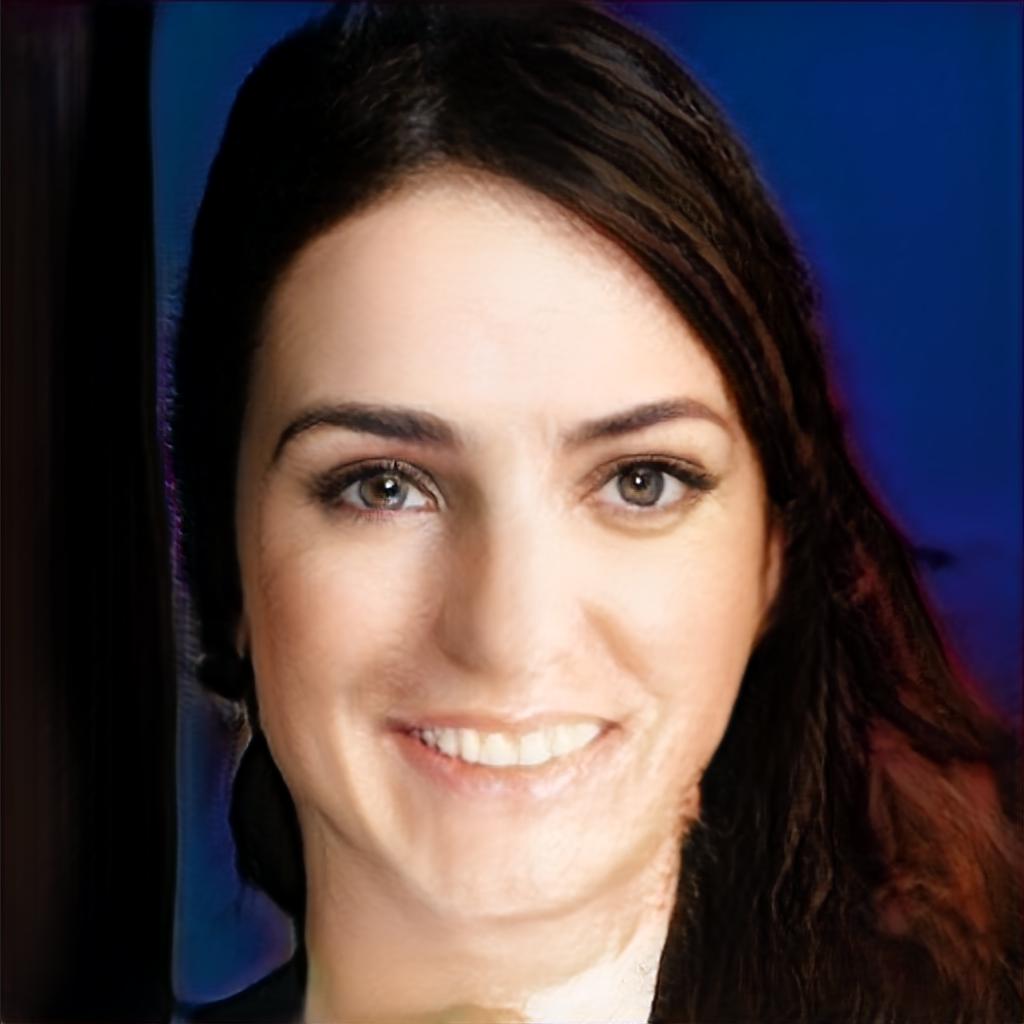}\\
\end{tabular}\\
{\footnotesize Generation inspired from \brtwo{images of faces}, model trained on CelebaHQ. \omitme{The target images were cropped around the faces, rotated so that the eyes lie on an horizontal line and scaled to make their sizes approximately equal.} \\}
\begin{tabular}{lcccc}
& & & & \\
\textbf{Target} & \textbf{2PDE}& \textbf{DOPO} & \textbf{Adam} & \textbf{LBFGS} \\
\includegraphics[width=0.18\linewidth]{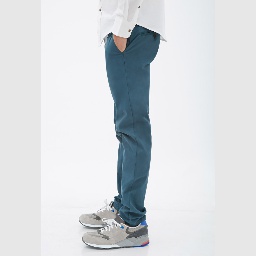}~~ &
\includegraphics[width=0.18\linewidth]{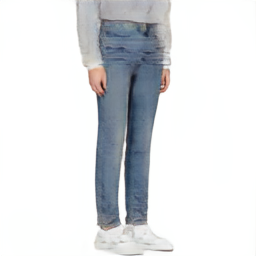} &
\includegraphics[width=0.18\linewidth]{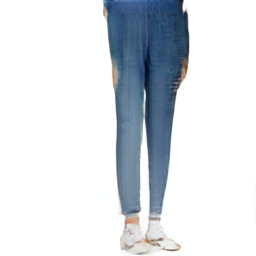}&
\includegraphics[width=0.18\linewidth]{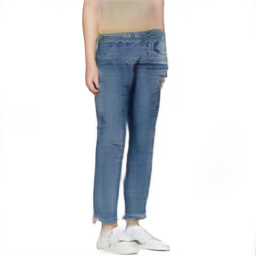}&
\includegraphics[width=0.18\linewidth]{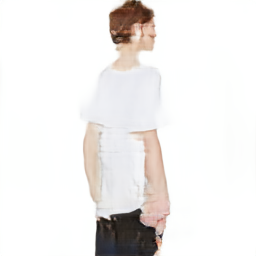}\\
\includegraphics[width=0.18\linewidth]{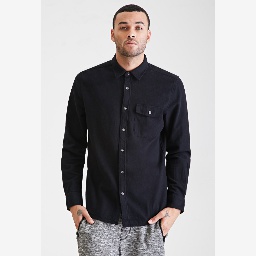}&
\includegraphics[width=0.18\linewidth]{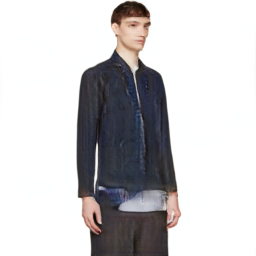} &
\includegraphics[width=0.18\linewidth]{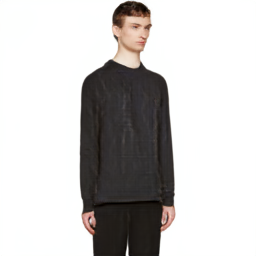}&
\includegraphics[width=0.18\linewidth]{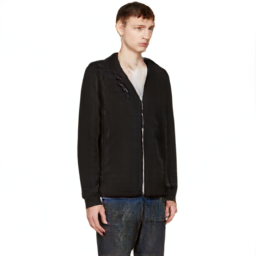}&
\includegraphics[width=0.18\linewidth]{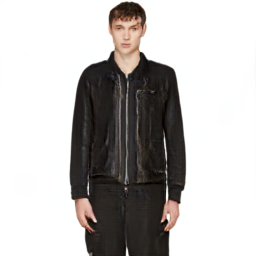}\\
\includegraphics[width=0.18\linewidth]{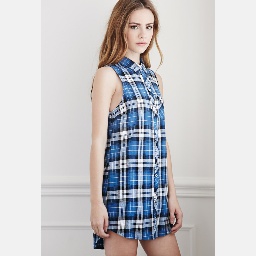}&
\includegraphics[width=0.18\linewidth]{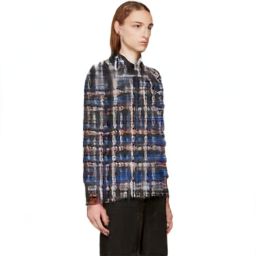}&
\includegraphics[width=0.18\linewidth]{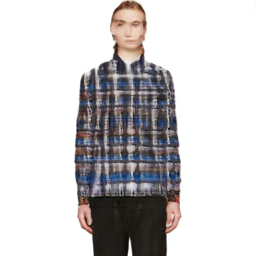}&
\includegraphics[width=0.18\linewidth]{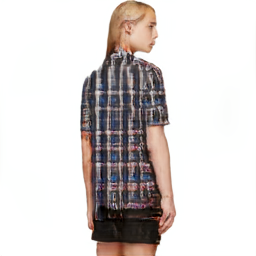}&
\includegraphics[width=0.18\linewidth]{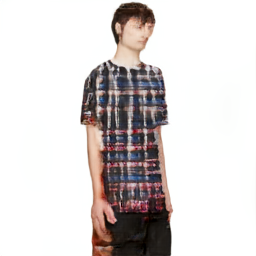}\\
\end{tabular}\\
{\footnotesize Generation inspired from DeepFashion, model trained on FashionGen.\\}
\caption{\ccc{Misspecified generations using VGG loss, budget of 3000 iterations. As confirmed by the human study results presented later, gradient-based approaches seem best for retrieving related faces, but are outperformed by derivative free approaches 2PDE and DOPO on fashion image retrieval.} \label{fig:Deep_fashion}}
\end{figure}

\begin{figure*}[htb]
\begin{tabular}{cc}
\includegraphics[width=0.48\linewidth]{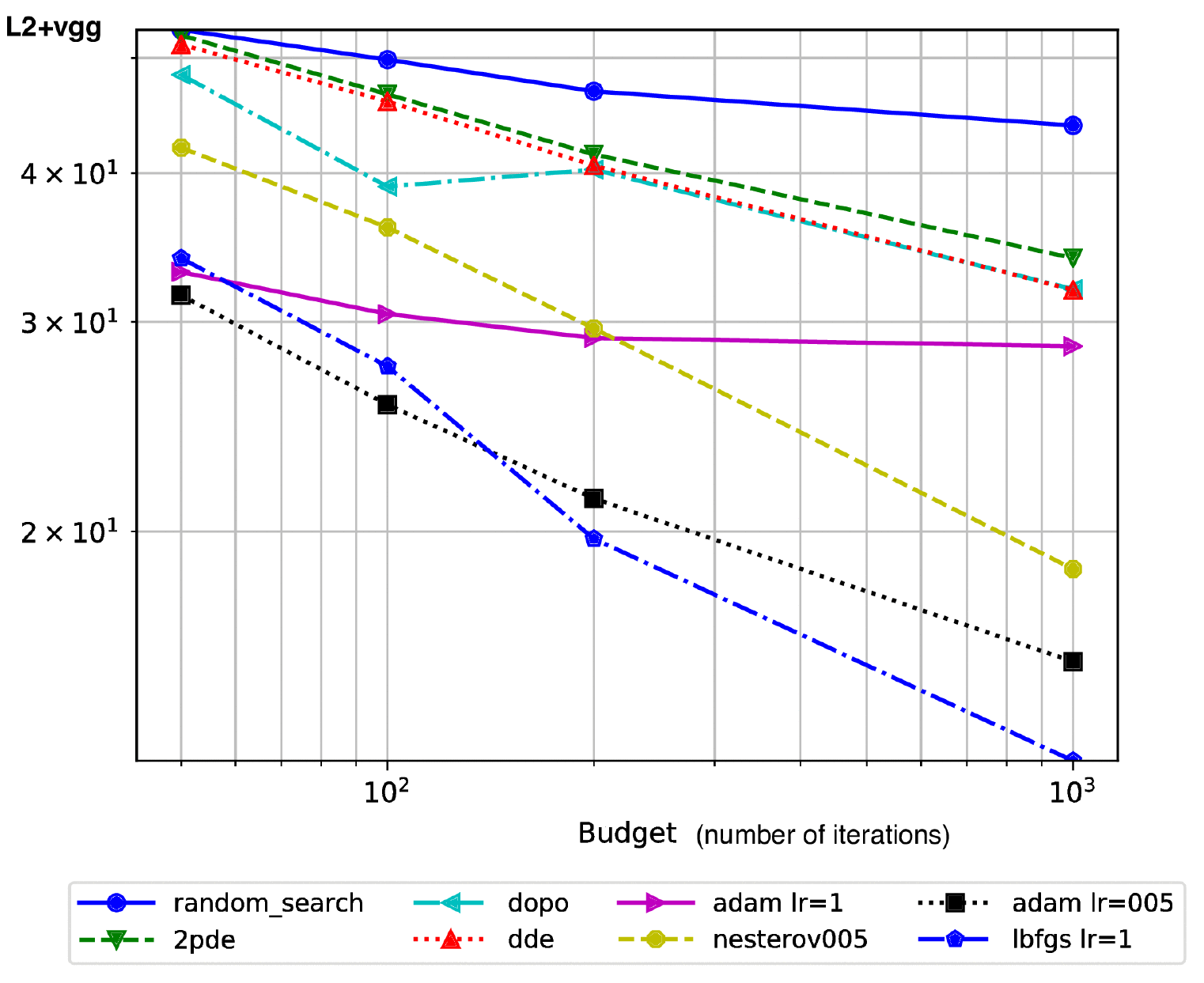}&
\includegraphics[width=0.48\linewidth]{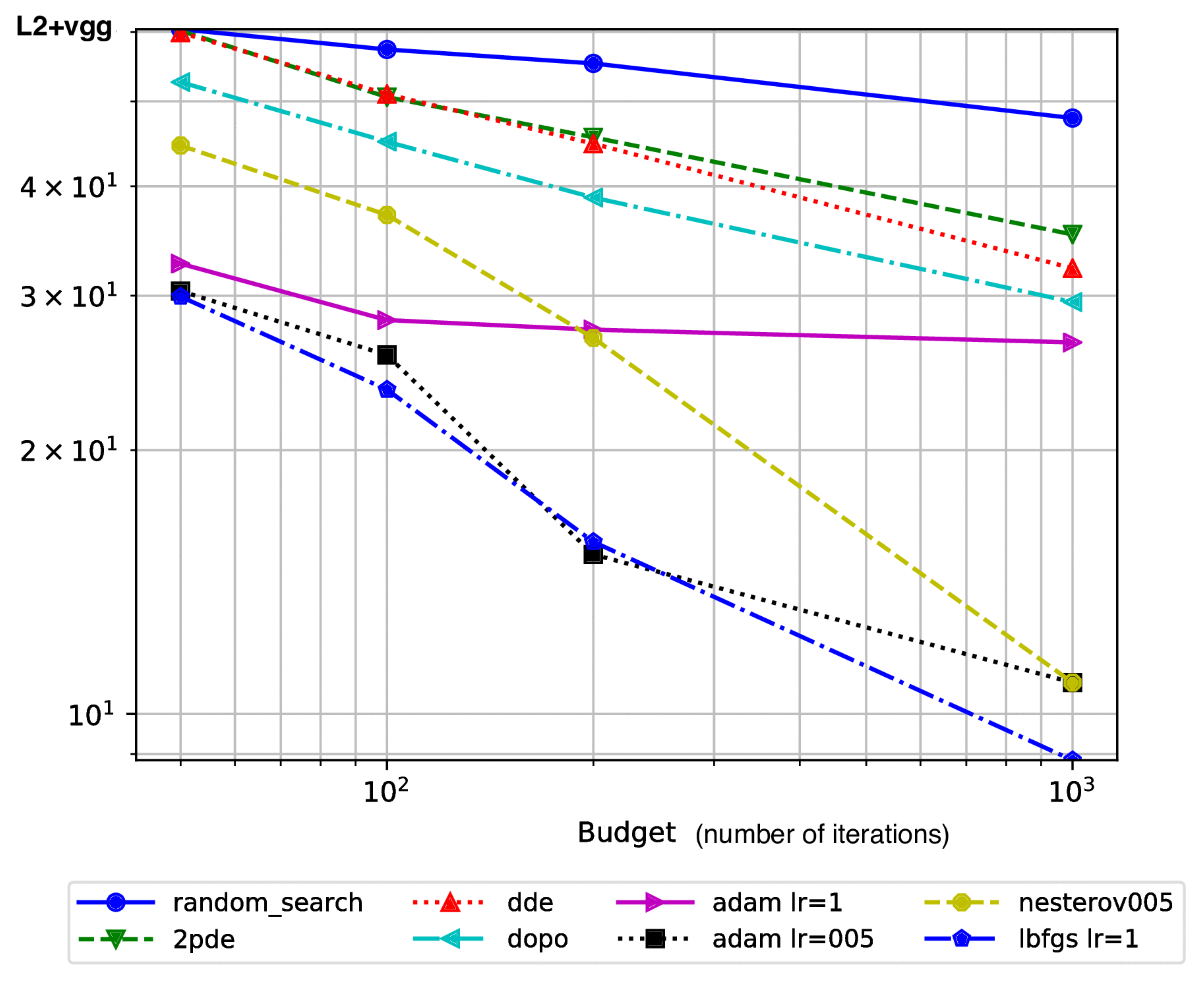}\\
(a) on the CelebA dataset & (b) On the DTD dataset \\
with criterion L2 + VGG. 
	& with criterion L2 + VGG without realism penalty.\\
\end{tabular}
	\caption{Retrieval performance measured with the criterion $\mathcal{C}$ (Reconstruction case). LBFGS dominates on both CelebA and DTD datasets. Here the x-axis is the number of calls to the generator. \cc{Gradient-based methods are approximately five times slower (per iteration) than comparison-based methods} The methods rank similarly when working with other training datasets. \TIPnewanswer{Random search is vastly outperformed by all other techniques.} \label{reconst}}
	
\label{fig:curves}
\end{figure*}
\begin{figure}[htb]
\center
\begin{tabular}{cccc}
Target & LBFGS & 2PDE & DOPO \\ 
\includegraphics[width=0.2\linewidth]{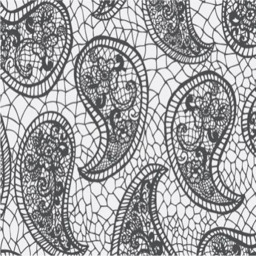}&
\includegraphics[width=0.2\linewidth]{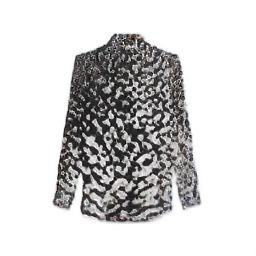}&
\includegraphics[width=0.2\linewidth]{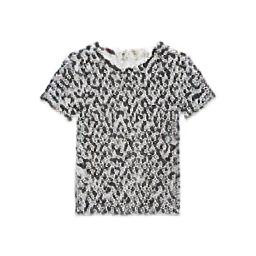}&
\includegraphics[width=0.2\linewidth]{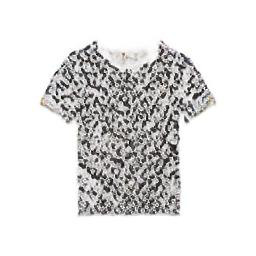}\\
\includegraphics[width=0.2\linewidth]{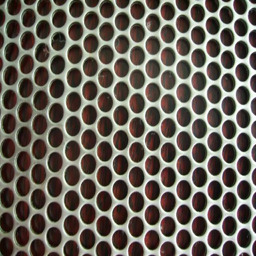}&
\includegraphics[width=0.2\linewidth]{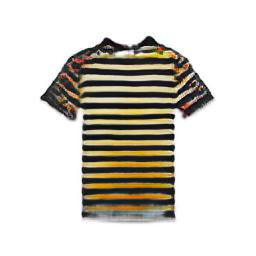}&
\includegraphics[width=0.2\linewidth]{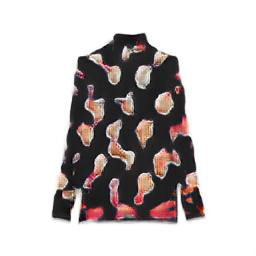}&
\includegraphics[width=0.2\linewidth]{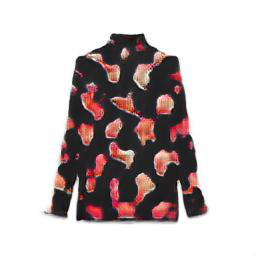}\\
\includegraphics[width=0.2\linewidth]{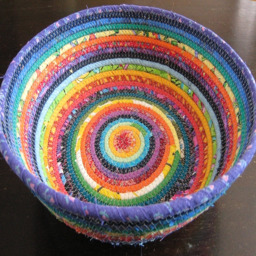}&
\includegraphics[width=0.2\linewidth]{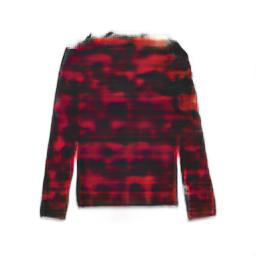}&
\includegraphics[width=0.2\linewidth]{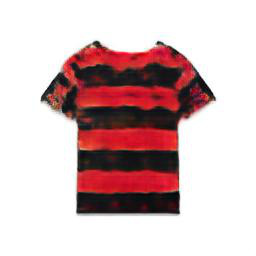}&
\includegraphics[width=0.2\linewidth]{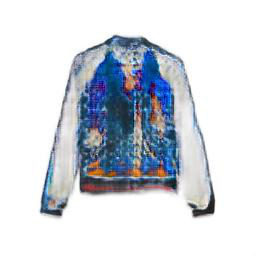}\\
\includegraphics[width=0.2\linewidth]{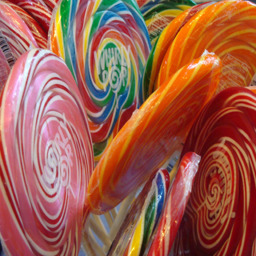}&
\includegraphics[width=0.2\linewidth]{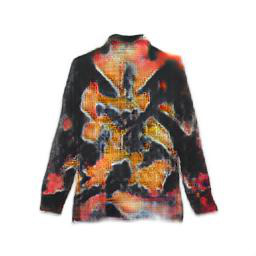}&
\includegraphics[width=0.2\linewidth]{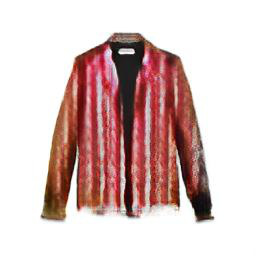}&
\includegraphics[width=0.2\linewidth]{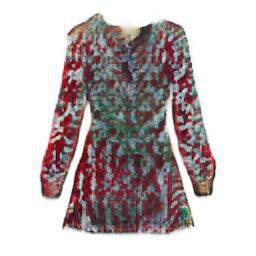}\\

\end{tabular}
\caption{Images from the RTW network retrieved from texture images (DTD) using the VGG loss.\label{fig:RTW}}
\end{figure}

 Figure \ref{fig:curves} presents a benchmark of the best performing algorithms we tested. 
Numerically, \ccc{as shown in Table \ref{misscases},} LBFGS is the best algorithm asymptotically in most cases. DOPO performed best among comparison-based methods, \ccc{and is} compatible with user preferences (i.e. when no criterion or no target image exists).
As far as the loss is concerned, all optimizers show similar behaviors on different datasets with different loss criteria. Numerically, random search is the worst performer, followed by gradient-free approaches (2PDE, DOPO, DDE), and finally gradient-based approaches.
\ccc{Qualitatively,} all optimization methods except Nesterov gave fair results with all training datasets (see Figures  \ref{fig:Deep_fashion} and \ref{fig:RTW}).
\ccc{In terms of computation time,} the time per iteration is essentially the same for all comparison-based algorithms on the one hand, and for all gradient-based methods on the other, because most of the computation time lies in the computation of $G(z)$ or $\nabla G(z)$. Gradient-based methods are 5 times slower than comparison-based methods due to the additional cost of computing $\nabla G(z)$.

Table \ref{res2} presents results of a human study conducted to evaluate the best optimizer for \ccc{retrieving generations from} GANs trained on different datasets. Surprisingly, LBFGS, which has the lowest criterion scores is outperformed by other approaches. On DTD, 2PDE ranks first, followed by DOPO. These methods rank similarly when using samples from DeepFashion on a dataset trained on FashionGen. We attribute this phenomenon to the tendency of evolutionary algorithms to prefer stable robust optima \cite{inoculation}. Meanwhile, Adam seems to be the best option for face retrieval. The generative function may be more regular. 
When working with a \ccc{class-conditioned} generator, setting the values of the discrete part of the latent vector resulted in more stable and visually more pleasant generations (See Table \ref{res2}a). This is held true regardless of the optimizer or the loss criterion considered.

\subsection{Human evolution(HEVOL) for preference based generation }\label{fc}

\subsubsection{Facial composite}

Facial composites were originally based on a reconstruction methodology based on questions and answers on individual facial features. \cite{deux,trois,solomon2009new} pointed out that holistic methods, based on global faces, are competitive.  
\ccc{Most systems \cite{six,sept,huit} do not include deep generators, and we extend \cite{tog} by increasing the number of latent variables, testing a wider ranger of evolutionary algorithms including an automatic update of the mutation rates, using higher quality generators and new datasets, resulting in a significantly better quality of obtained images.}
\omitme{\ccc{Instead}, we focus on a global facial composite system. In our experiments,} We worked with HEVOL combined with a generator trained on CelebaHQ.
\omitme{We use, as \cite{deux,trois,quatre}, an evolutionary algorithm, combined with a generator trained on CelebaHQ.}

\begin{figure}
\center
\begin{tabular}{ccc}
\textbf{Target} &\textbf{HEVOL} & \textbf{Random Search}\\
\includegraphics[height=.3\linewidth]{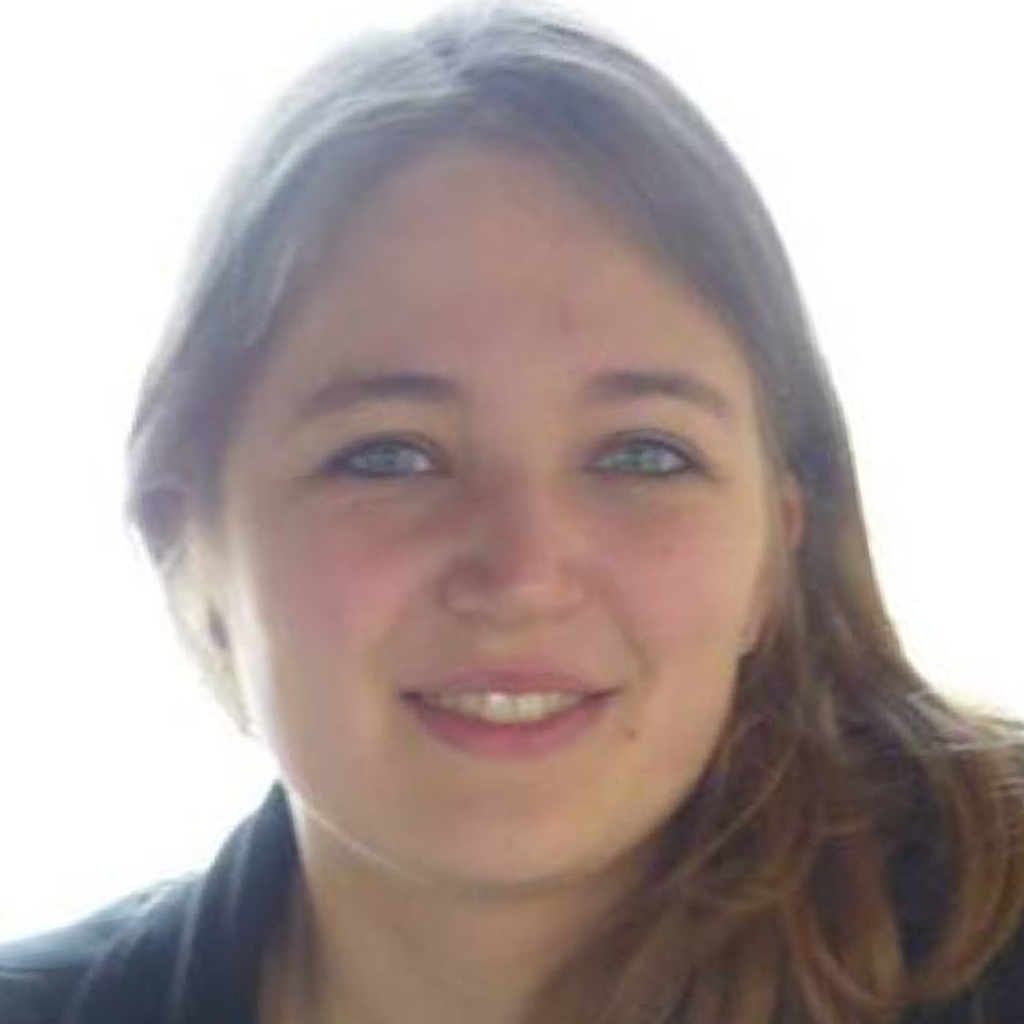} &
\includegraphics[height=.3\linewidth]{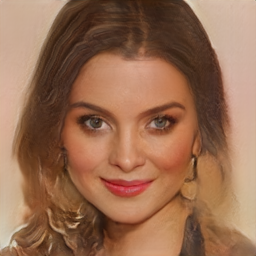}  & 
\includegraphics[height=.3\linewidth]{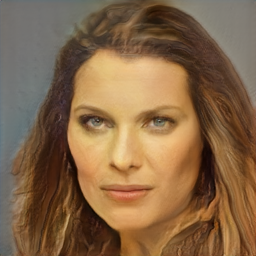}\\
\includegraphics[height=.3\linewidth]{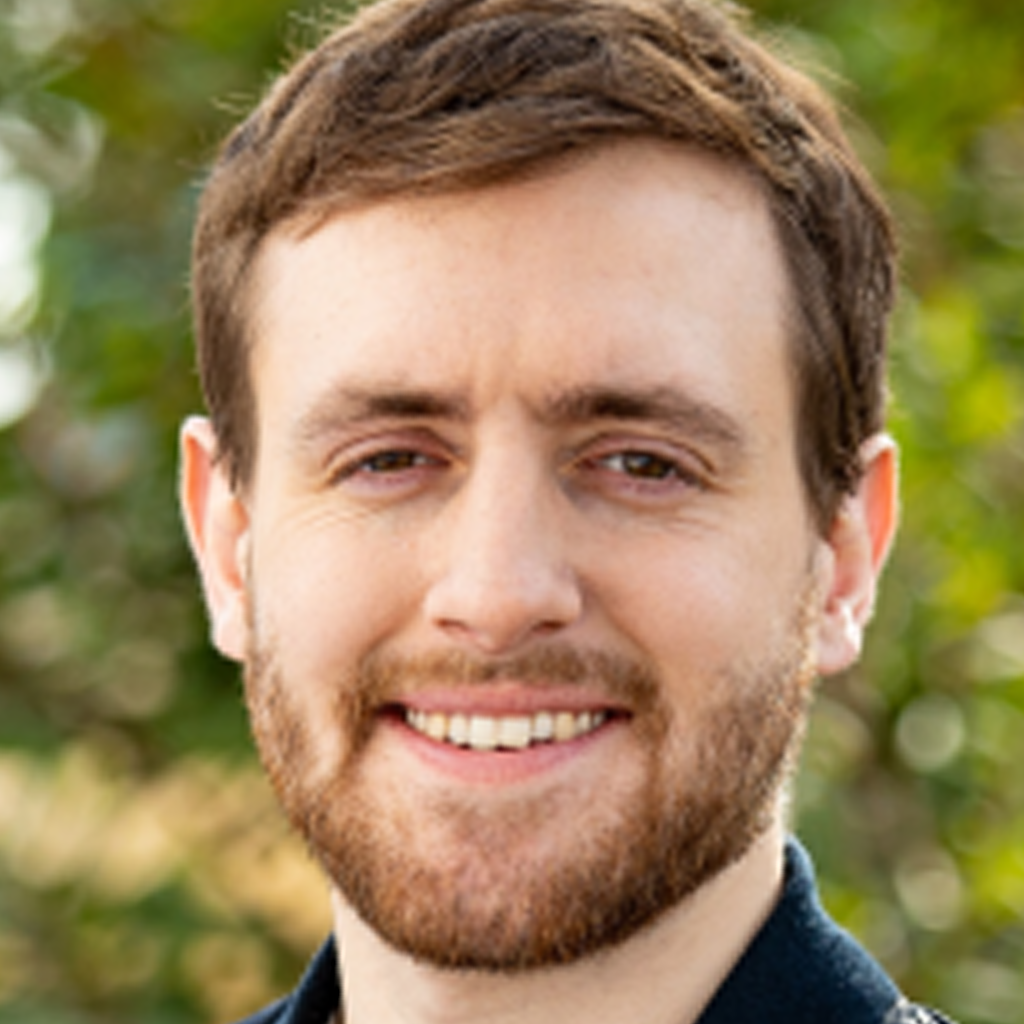} &
\includegraphics[height=.3\linewidth]{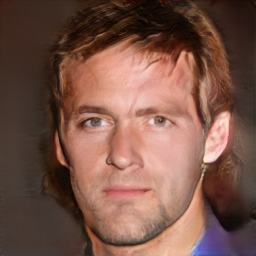}  & 
\includegraphics[height=.3\linewidth]{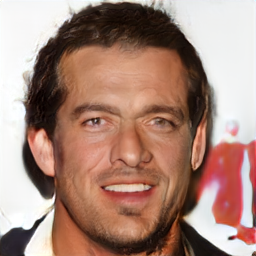}
\end{tabular}
	\caption{\label{jk}A \brtwo{real human face picture} (left) and its reconstructions obtained with HEVOL (middle) and Random Search (right) after 6 iterations (less than 3 minutes of user time).}
\end{figure}

We use a random \ccc{face picture} (Figure \ref{jk}, left) as a target for our facial recognition experiment. This is a misspecified case as the target's face is not in the CelebaHQ dataset. At each iteration, the user was asked to select $\mu=5$ favorite  among $1+\lambda=28$ images. We found that we could get better performance and convenience if we allowed the user to select the same image several times. We obtained convincing outputs in as few as six iterations \br{(see Figure \ref{jk})}.

\ccc{To validate the stronger performance of HEVOL compared to a random search, we conducted the following human study.}
We generated faces using 30 different target images \omitme{from FFHQ with both}\ccc{using either} HEVOL or a random search method, where the user selects the best of 23 images generated with Gaussian latent vectors at each iteration.
\br{Human annotators preferred the image generated by HEVOL to that generated by random search for 73.3\% (Confidence Interval [54.1\%, 87.7\%]) of the 30 images. \TIPnewanswer{They preferred the image generated by random search for only 16.7\% of the images and had no preference for 10.0\% of them: HEVOL performs better than random search.} }

\begin{figure}
\center
\begin{tabular}{cc}
\includegraphics[height=.45\linewidth]{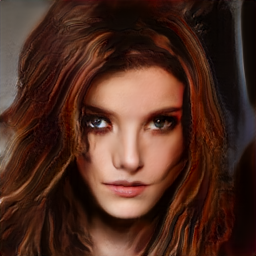}  & 
\includegraphics[height=.45\linewidth]{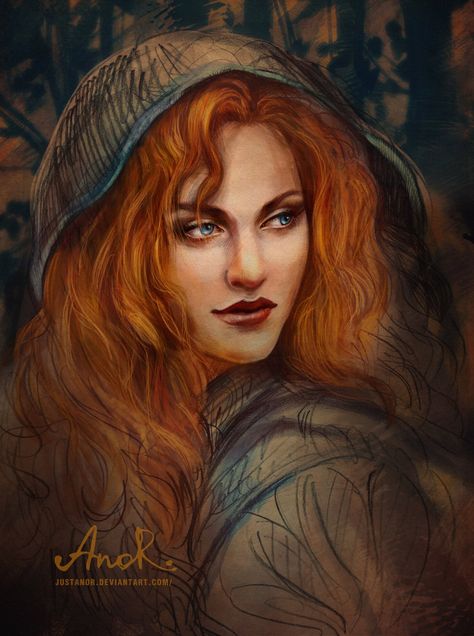}
\end{tabular}
\begin{tabular}{cccc}
     \includegraphics[height=0.2\linewidth]{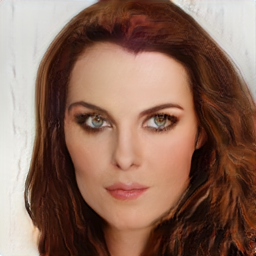}& \includegraphics[height=0.2\linewidth]{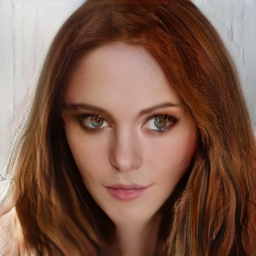}  &\includegraphics[height=0.2\linewidth]{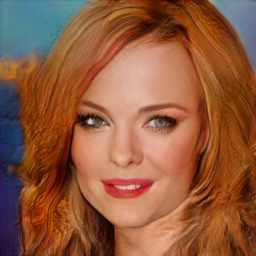} &\includegraphics[height=0.2\linewidth]{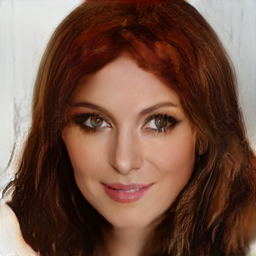}
\end{tabular}
	\caption{\br{\label{fig:image_from_description}On the top left, an image generated with HEVOL using the description of Triss Merigold from the book ``the Witcher" and a GAN trained on CelebA-HQ. On the top right, a drawing from \ccc{an artist (}AnoR) representing the same character and which happens to look similar to the image we generated. The human selected the pictures that fit the description ``A beautiful woman with blue eyes and chestnut hair with sheen of red" and without using any target image. On the bottom, four other images generated with HEVOL to fit the same description.}}
\end{figure}

The main advantage of HEVOL is that it allows the generation of images without a target image. For instance, one can generate a picture from the description of a character from a book (see Figure \ref{fig:image_from_description}), a police sketch from memory or a piece of clothing corresponding to a given season or style (see Figure \ref{figimg}).
This budget \ccc{of about 140 images} is far too low for any non human\ccc{-based} algorithm. LBFGS fails to reach this quality even with 400 iterations. \TIPanswer{Generating a single image using HEVOL generally takes about as much time as using LBFGS, though with much more variance. The time taken to generate an image using HEVOL is highly dependent on the human speed using the algorithm. For LBFGS on facial composites generation, each iteration takes about 0.043 seconds and the images generated in Figure~\ref{fig:reconstruction1} and \ref{fig:Deep_fashion} for instance were generated in about two minutes. The time taken to choose the five best images for human annotators is  variable.\omitme{It depends on the human and on the iteration:} Choosing the best images is much easier at the beginning when most images are not similar to the target. Humans are able to complete six iterations in less than three minutes, with some runs taking less than a minute for the images generated for the human study and Figure~\ref{jk}.}

%----------------------------------------------------------------------

\subsubsection{\TIPanswer{Text2img flower images retrieval}}\label{sec:flower_HEVOL}
\begin{figure}
\center
\begin{tabular}{ccc}
\textbf{Target} &\textbf{HDGAN} & \textbf{HEVOL}\\
\begin{minipage}{.2\linewidth}\vspace*{-2.5cm}
\includegraphics[height=.95\linewidth]{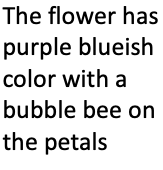}  \end{minipage} & 
\includegraphics[height=.25\linewidth]{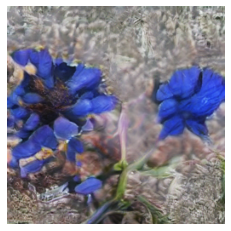}&
\includegraphics[height=.25\linewidth]{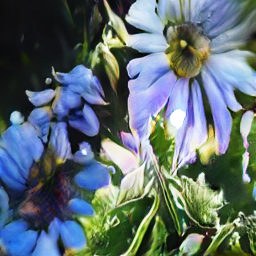}\\
\begin{minipage}{.2\linewidth}\vspace*{-2.5cm}
 \includegraphics[height=.95\linewidth]{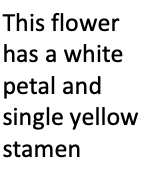}\end{minipage}& 
 \includegraphics[height=.25\linewidth]{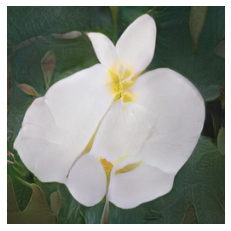}& 
 \includegraphics[height=.25\linewidth]{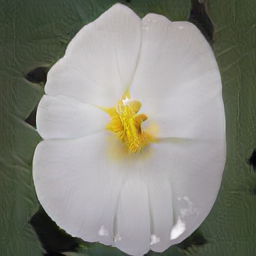}\\
 \begin{minipage}{.2\linewidth}\vspace*{-2.5cm}
\includegraphics[height=.95\linewidth]{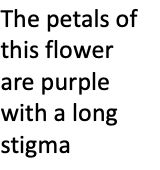}\end{minipage} &
\includegraphics[height=.25\linewidth]{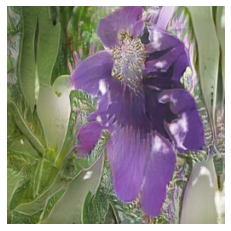} & 
\includegraphics[height=.25\linewidth]{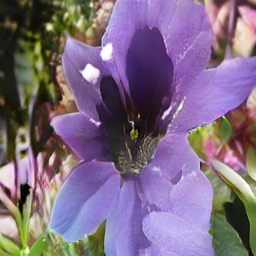}
\end{tabular}
	\caption{\label{fig:HEVOL_flowers} \TIPanswer{Flower images generated using the original HDGAN and HEVOL with the same GAN for given flower descriptions. With HEVOL, we are often able to get colors that match the description better or to generate more realistic flower images. }}
	\label{fig:flowers}
\end{figure}

\TIPanswer{HEVOL can also be used to retrieve images based on a text description. In that case, we compare the images obtained with the HDGAN from \cite{zhang2018photographic} for flower images generation from text descriptions using the Oxford-102 Flowers dataset \cite{nilsback2008automated} to those we obtain using HEVOL on the same GAN network.
HEVOL uses at most eight iterations (the user can stop early if satisfied), choosing five images out of 25 at each iteration. 
We ran a human study with seven participants who were asked to select the best results among 14 image pairs. Each pair contains an image obtained from HDGAN and an image obtained from HEVOL using HDGAN as a base algorithm.  
Obtained average human ratings are $77.6\pm 5.4\%$ in favor of HEVOL, confirming that HEVOL can be applied on text2img as well. Examples of results are displayed in Fig.~\ref{fig:flowers}. \TIPnewanswer{
Our approach can be successfully combined with Txt2Image, so that a few clicks by a user can significantly improve the generated image.}}

\subsubsection{Fashion image retrieval}\label{fir}
\begin{figure}
    \centering
    \begin{tabular}{cccc}
    \includegraphics[width=0.18\linewidth]{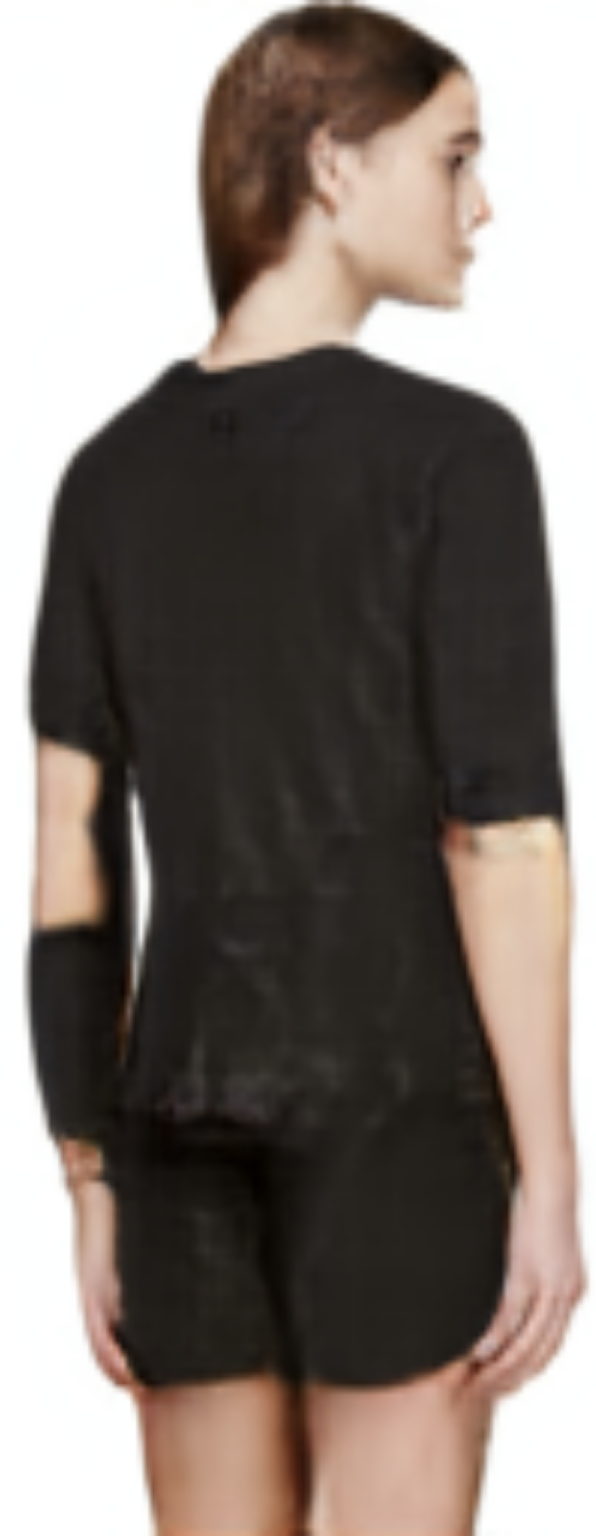}&
    \includegraphics[width=.2\linewidth]{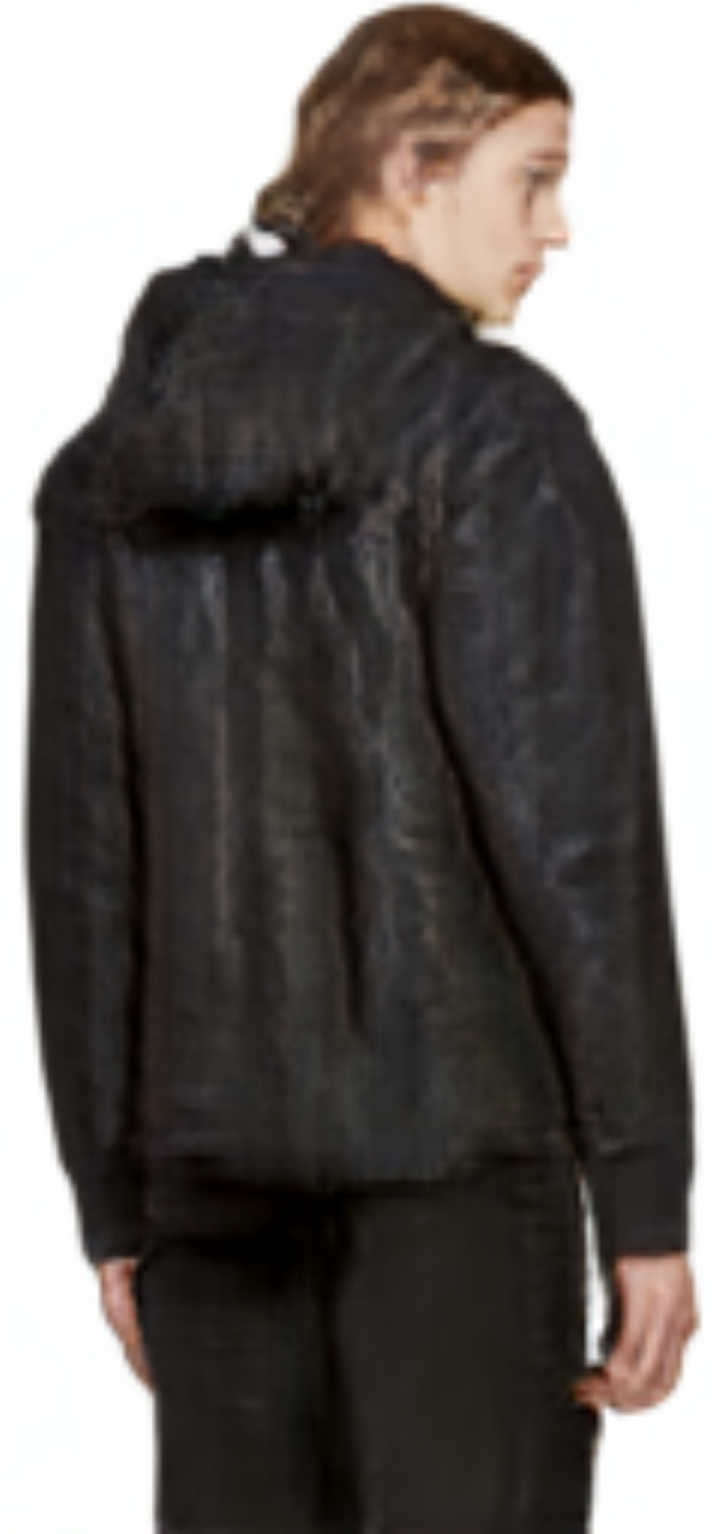}&
    \includegraphics[width=.15\linewidth]{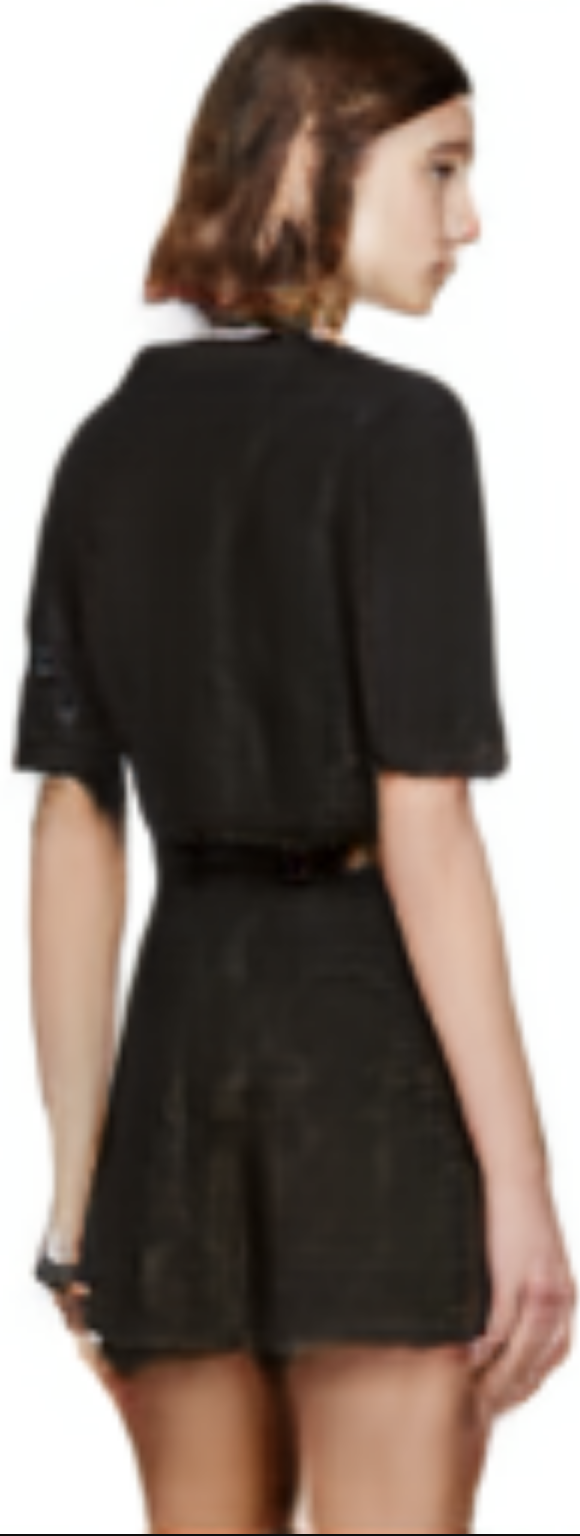}&
    \includegraphics[width=.2\linewidth]{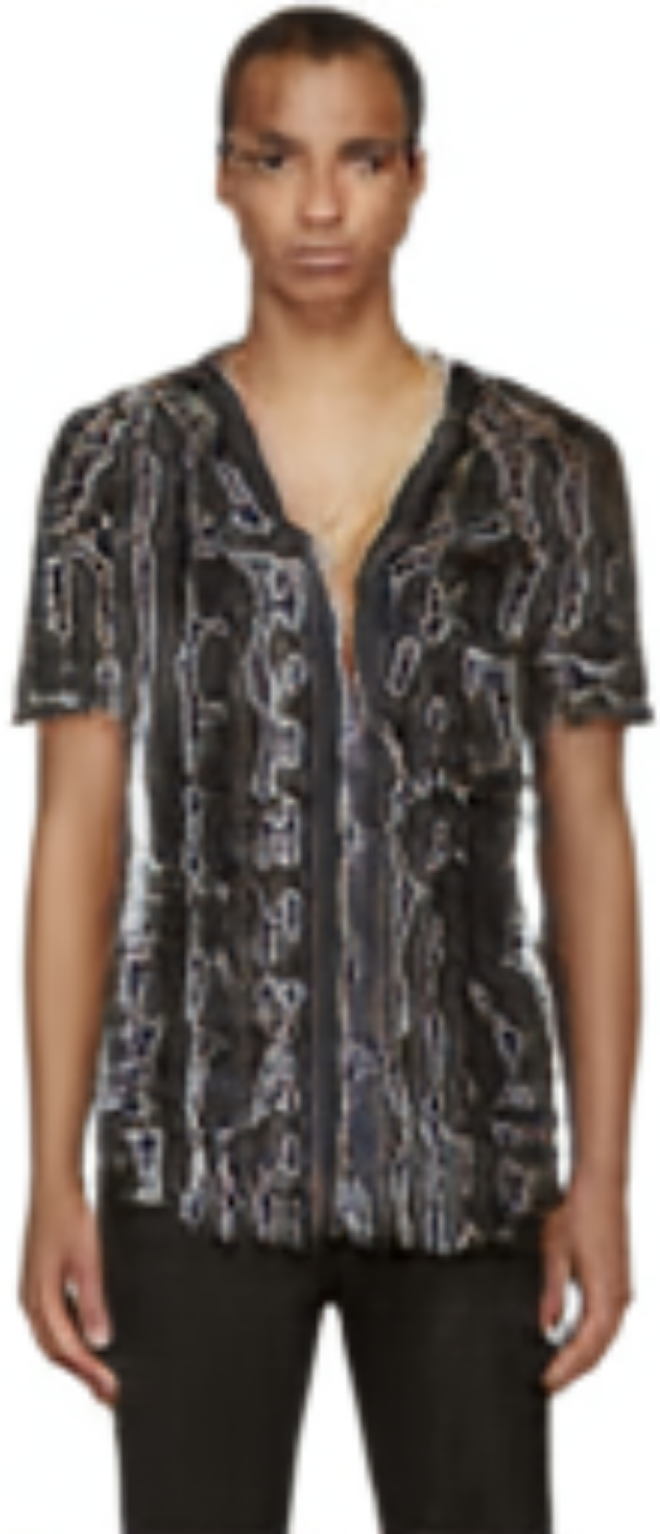}\\
    \end{tabular}\\
    %\includegraphics[width=1\linewidth]{IJCV/fg/redish}
 %%   \\
	\caption{User-chosen images obtained with HEVOL. The instructions were respectively to produce ``sportswear'', ``clothes for cold weather'', ``light clothes'', ``sophisticated''. 61 images were generated in each case, i.e. 4 generations of 15 images plus the initial one. 
	\label{figimg}}
\end{figure}

\ccc{HEVOL} can be used for fashion generation. \br{In this case, the user selects the images which are the best match for the piece of clothing they have in mind. }
We use this approach on a model trained on FashionGen and work with four batches of 16 images. Results are presented in Figure \ref{figimg}. Contrary to facial recognition, averaging performs poorly, so the user only selects one picture at each iteration.

%----------------------------------------------------------------------

\begin{table}
\begin{center}
{\small
\begin{tabular}{|cc|cccc|}
\hline
 && LBFGS  & Adam & DOPO & 2PDE  \\ 
 \hline
\multicolumn{1}{|c|}{DTD,} &cond & 14.4 & 9.6& 13.8 & \bf 25.3   \\
\multicolumn{1}{|c|}{targets} &no cond & 3.1 & 10.5 & \bf 12.6 & 10.7   \\
\multicolumn{1}{|c|}{from DTD} & sum &  17.5 & 20.1 & 26.4 & \bf 36.0\\
\hline
\multicolumn{2}{|c|}{CelebA-HQ ,}&&&&\\
\multicolumn{2}{|c|}{ targets from }& 29.2 & \bf 37.2  & 15.6  & 18.0 \\
\multicolumn{2}{|c|}{ \brtwo{random face pictures}} &&&&\\
\hline
\multicolumn{2}{|c|}{FashionGen, }&&&&\\
\multicolumn{2}{|c|}{targets from } & 11.7 & 20.9 & \bf 41.2 & 26.3 \\
\multicolumn{2}{|c|}{DeepFashion} &&&&\\
\hline
\end{tabular}}\\
\end{center}
\caption{Comparison of different retrieval algorithms on various datasets  (\% of retrieved images judged most similar to the target). DTD and FashionGen generations retrieved by evolutionary algorithms are found more related to targets, whereas CelebA-HQ generations work better with Adam. The Celeba-HQ case is easier to optimize - difficult cases lead to better results with evolutionary algorithms. It is consistent with the robustness qualities of these methods.}
\vspace{-0.5cm}
\label{res2}
\end{table}

\section{Conclusion}
\def\removethat{We generate images biased by similarity towards a target image or biased by humans preferences, for various datasets from faces (facial composites), textures and fashion generation. Our contributions include extending the combination of evolution and GAN\cite{tog} for interactive generation of images \br{of}\omitme{to} faces (Figs. \ref{todofigures}), textures, and fashion; combining VGG similarity and discriminator realism criterion into  \cite{todocitations}; doing such generations without human in the loop based on an inspirational target image (Figs. \ref{todofigures}); shown the unexpected performance of evolutionary algorithms in that context (discussion below); include validations by human raters (Figure \ref{todofigures}).

}

\otc{Searching the latent space becomes a classical issue in GANs, including applications in inpainting, facial composites and super-resolution.}
We show several ways to control the output of a trained generative model\ccc{, either by providing an inspirational image as input, or driving the search according to user preferences on the global generation.}\omitme{, extending \cite{baubau,imtoim,tog,Yeh_2017_CVPR,invert,srinspir}.} \ccc{When trained on \br{images of faces}\omitme{faces images,} w}ith a simple L2 loss and LBFGS, a GAN can build a rough approximation of almost any image and retrieve its own generations. We show the excellent numerical performance of gradient-based methods and in particular LBFGS to recover optima corresponding to existing generations. 
\otc{This result can be applied to various latent space searching methods.}
\br{Surprisingly, although they fail numerically compared to LBFGS, evolutionary methods like DOPO and 2PDE were more successful than gradient-based methods according to human opinion. This is particularly true for misspecified cases \ccc{(looking for a generation close to a target absent from training data)}.}
With a human user in the loop, the classical $(\mu/ \mu +\lambda)$ evolution strategy provides \omitme{in a few minutes reasonable} facial composites and \ccc{fashion image retrievals} in a few minutes. 
\br{Images can be generated from any arbitrary criterion and without using a target image}.
\ccc{We achieve unprecedented diverse high quality images, extending the combination of
evolution and GANs for interactive generation of images of
faces, textures, and fashion.} 
Our code is open sourced, and our results are validated by double-blind human rating.

\begin{figure}
\vspace{-0.5cm}
\center
\setlength\tabcolsep{1.5pt}
\begin{tabular}{ccc|ccc}
\textbf{Target} & \textbf{Best}& \textbf{Other} & \textbf{Target} & \textbf{Best}& \textbf{Other} \\
\includegraphics[width=0.15\linewidth,height=0.15\linewidth]{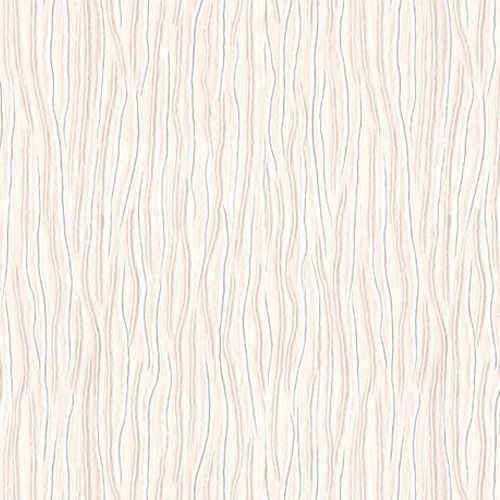} &
\includegraphics[width=0.15\linewidth]{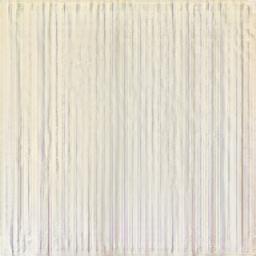} &
\includegraphics[width=0.15\linewidth]{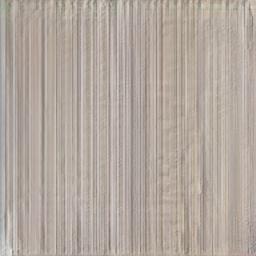}~ &
~\includegraphics[width=0.15\linewidth,height=0.15\linewidth]{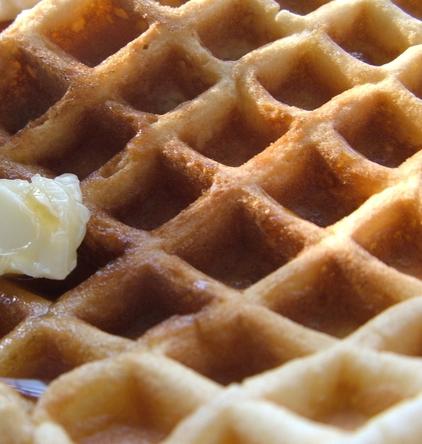} &
\includegraphics[width=0.15\linewidth]{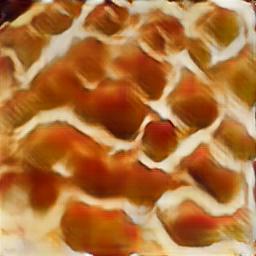} &
\includegraphics[width=0.15\linewidth]{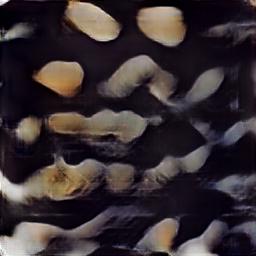} \\
&LBFGS&2PDE&&2PDE&LBFGS\\
\includegraphics[width=0.15\linewidth,height=0.15\linewidth]{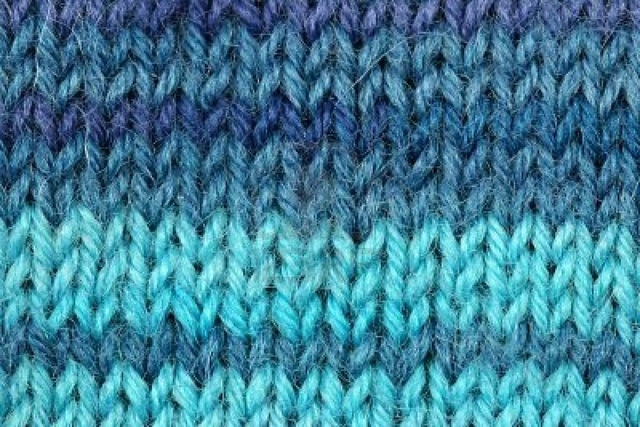} &
\includegraphics[width=0.15\linewidth]{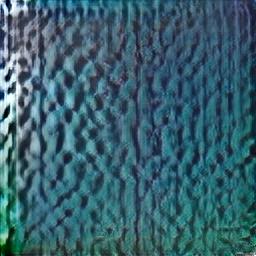} &
\includegraphics[width=0.15\linewidth]{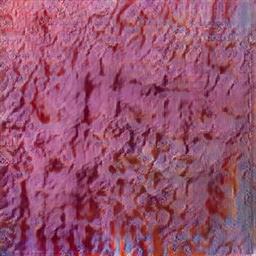}~ &
~\includegraphics[width=0.15\linewidth,height=0.15\linewidth]{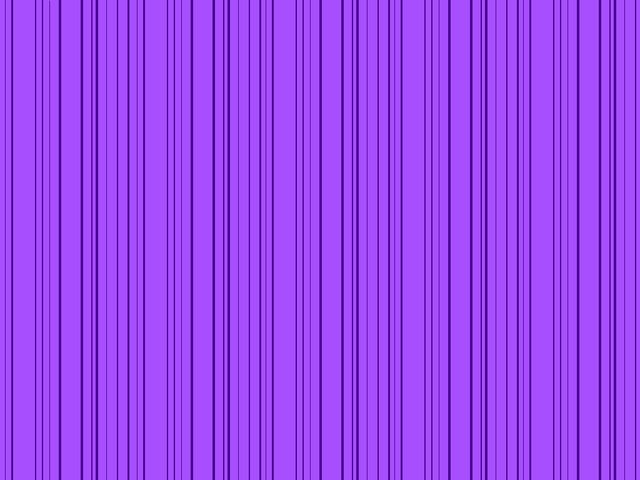} &
\includegraphics[width=0.15\linewidth]{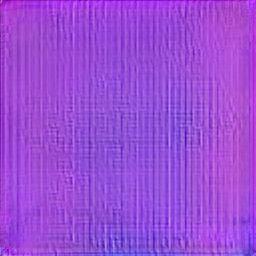} &
\includegraphics[width=0.15\linewidth]{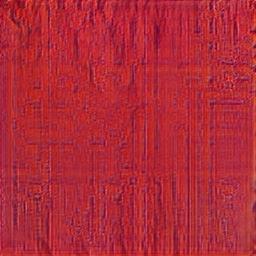} \\
&LBFGS&2PDE &&Adam&2PDE\\
\includegraphics[width=0.15\linewidth,height=0.15\linewidth]{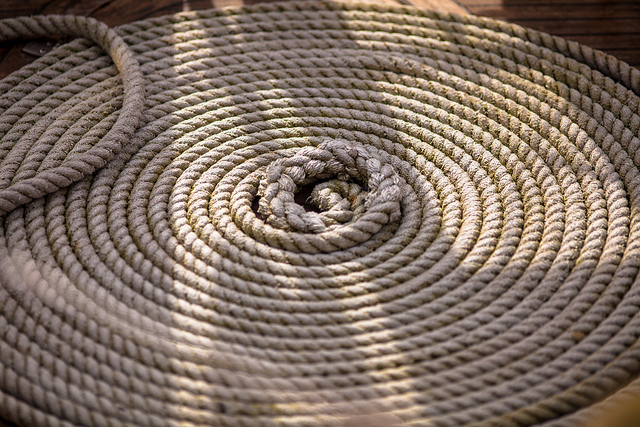} &
\includegraphics[width=0.15\linewidth]{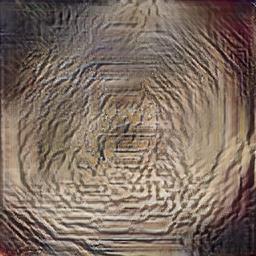} &
\includegraphics[width=0.15\linewidth]{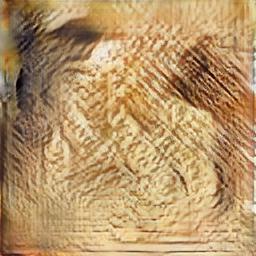}~ &
~\includegraphics[width=0.15\linewidth,height=0.15\linewidth]{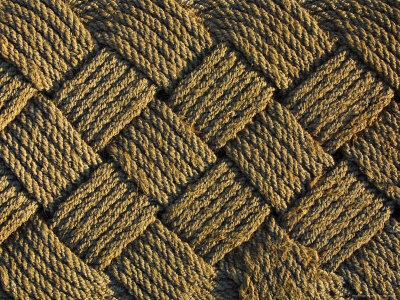} &
\includegraphics[width=0.15\linewidth]{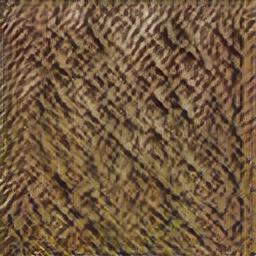} &
\includegraphics[width=0.15\linewidth]{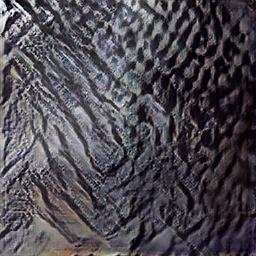} \\
&2PDE&LBFGS && 2PDE&LBFGS\\
\end{tabular}\\
\caption{\label{fig:FashionAgreements} Best retrieved DTD images as judged by humans. Similarity criterion: VGG loss. \ccc{On these texture images, 2PDE is best, especially for complex patterns, followed by LBFGS.}
\vspace{-0.5cm}} 
\end{figure}

\section*{Acknowledgment}
We thank Danielle Rothermel, Natalia Neverova, Matthijs Douze, Arthur Szlam, Diane Bouchacourt and Louis Martin for their useful input to this work.

\bibliographystyle{plain}      
\bibliography{biblio}  

\appendix
\paragraph{Layers considered in vgg19}
When using a VGG-19 network as a feature extractor, we always consider the output of 3 ReLU layers:
\begin{itemize}
    \item The first one after the second MaxPooling operation
    \item The first one after the third MaxPooling operation
    \item The first one after the fourth MaxPooling operation
\end{itemize}

\paragraph{Hardware details}
The progressive GAN networks were trained on two NVIDIA Tesla V100-SXM2 and took less than a week per model. We used the same hardware for the inspired generation. Our program relies on the Pytorch python framework.

\paragraph{Scaling}
When judging the similarity between a pair of images, we are more interested in the main patterns of the image than in the tiniest details. That is why both $I$ and $G(z)$ are resized to $128\times128$ before being fed to the feature network \cc{(VGG layers as described above)}. Therefore, both images can be of arbitrary size and a high resolution generation can be made out of a lower resolution reference.
\brtwo{For facial composites generation, we ensure that the face is centered and the eyes are on a horizontal line. We also rescale the images so that the faces all have approximately the sime size.}

\end{document}